\documentclass{article}

\PassOptionsToPackage{numbers, compress}{natbib}

\usepackage[preprint]{neurips_2023}

\usepackage[utf8]{inputenc} 
\usepackage[T1]{fontenc}    
\usepackage{url}            
\usepackage{booktabs}       
\usepackage{amsfonts}       
\usepackage{nicefrac}       
\usepackage{microtype}      
\usepackage{xcolor}         
\usepackage{xspace}
\usepackage{amsmath,amssymb,amsfonts,dsfont,pifont,bm,bbm,mathrsfs,mathtools,nicefrac}
\usepackage{algorithm,algpseudocode,listings}
\usepackage{booktabs,multirow,adjustbox,diagbox,threeparttable}
\definecolor{citeblue}{RGB}{48,111,186}
\usepackage[pagebackref=false,breaklinks=true,colorlinks=true,citecolor=citeblue,bookmarks=false]{hyperref}
\usepackage{cleveref}  
\usepackage{wrapfig}
\usepackage{subfig}

\crefname{section}{Sec.}{Secs.}
\Crefname{section}{Section}{Sections}
\crefname{table}{Tab.}{Tabs.}
\Crefname{table}{Table}{Tables}
\crefname{figure}{Fig.}{Figs.}
\Crefname{figure}{Figure}{Figures}
\crefname{equation}{Eq.}{Eqs.}
\Crefname{equation}{Equation}{Equations}
\crefname{appendix}{Appendix}{Appendix}
\usepackage{color, colortbl}
\definecolor{LightGray}{gray}{0.9}

\newcommand{\method}{LSV-GAN\xspace}

\newcommand{\fashion}{\textsc{DeepFashion}\xspace}
\newcommand{\shhq}{\textsc{SHHQ}\xspace}
\newcommand{\aist}{\textsc{AIST++}\xspace}

\usepackage{multirow}
\usepackage{array}

\newcommand{\moniker}{LSVs}

\newcommand{\smpl}{M}
\newcommand{\pose}{\boldsymbol{\theta}}
\newcommand{\shape}{\boldsymbol{\beta}}
\newcommand{\mesh}{\mathsf{M}}
\newcommand{\verts}{\mathsf{V}}
\newcommand{\faces}{\mathsf{F}}
\newcommand{\R}{\mathbb{R}}
\DeclareMathOperator{\lbs}{LBS}
\DeclareMathOperator{\uv}{UV}

\newcommand{\jointregressor}{J}
\newcommand{\texture}{\mathsf{T}}
\newcommand{\texLayers}{\mathcal{T}}
\newcommand{\vcolor}{\mathbf{c}}
\renewcommand{\v}{\mathbf{v}}
\newcommand{\vn}{\mathbf{n}}
\newcommand{\nMesh}{N}
\newcommand{\density}{o}
\newcommand{\colorSeq}{\mathbf{C}}
\newcommand{\densitySeq}{\mathbf{O}}
\newcommand{\pix}{\mathbf{p}}
\newcommand{\weight}{w}

\usepackage{xpunctuate}

\providecommand{\ie}[0]{i.e\xperiod}

\usepackage{array}
\newcolumntype{L}[1]{>{\raggedright\let\newline\\\arraybackslash\hspace{0pt}}m{#1}}
\newcolumntype{C}[1]{>{\centering\let\newline\\\arraybackslash\hspace{0pt}}m{#1}}
\newcolumntype{R}[1]{>{\raggedleft\let\newline\\\arraybackslash\hspace{0pt}}m{#1}}

\title{Efficient 3D Articulated Human Generation with Layered Surface Volumes}
\author{
  Yinghao Xu$^{1,2}$\thanks{work done when Yinghao was a visiting student at Stanford University}\\
	\texttt{yhxu@stanford.edu}\\
    \texttt{xy119@ie.cuhk.edu.hk}\\
	\And
   Wang Yifan$^1$\\
	\texttt{yifan.wang@stanford.edu}\\
    \And
  Alexander W. Bergman$^1$ \\
	\texttt{awb@stanford.edu}\\
	\And
  Menglei Chai$^3$\\
	\texttt{mengleichai@google.com}\\
	\And
  Bolei Zhou$^4$\\
	\texttt{bolei@cs.ucla.edu}\\
	\And
  Gordon Wetzstein$^1$\\
	\texttt{gordonwz@stanford.edu}\\
 \And
     $^1$Stanford \qquad
    $^2$CUHK \qquad
    $^3$Google \qquad
    $^4$UCLA
}

\begin{document}

\maketitle

\vbox{%
	\vskip -0.15in
	\hsize\textwidth
	\linewidth\hsize
	\centering
	\normalsize
	\tt\href{https://www.computationalimaging.org/publications/lsv/}{computationalimaging.org/publications/lsv/}
	\vskip 0.28in
}

\begin{abstract}
Access to high-quality and diverse 3D articulated digital human assets is crucial in various applications, ranging from virtual reality to social platforms. Generative approaches, such as 3D generative adversarial networks (GANs), are rapidly replacing laborious manual content creation tools. However, existing 3D GAN frameworks typically rely on scene representations that leverage either template meshes, which are fast but offer limited quality, or volumes, which offer high capacity but are slow to render, thereby limiting the 3D fidelity in GAN settings. In this work, we introduce layered surface volumes (LSVs) as a new 3D object representation for articulated digital humans. LSVs represent a human body using multiple textured mesh layers around a conventional template. These layers are rendered using alpha compositing with fast differentiable rasterization, and they can be interpreted as a volumetric representation that allocates its capacity to a manifold of finite thickness around the template. Unlike conventional single-layer templates that struggle with representing fine off-surface details like hair or accessories, our surface volumes naturally capture such details. LSVs can be articulated, and they exhibit exceptional efficiency in GAN settings, where a 2D generator learns to synthesize the RGBA textures for the individual layers. Trained on unstructured, single-view 2D image datasets, our LSV-GAN generates high-quality and view-consistent 3D articulated digital humans without the need for view-inconsistent 2D upsampling networks. 
\end{abstract}
\section{Introduction} 
\label{sec:intro}

High-quality 3D articulated digital human assets are becoming increasingly important for several industries, such as gaming, VR/AR, and social platforms. Manual authoring of these assets, however, is a laborious task that requires domain expertise and artistic skills. By automating the asset generation, generative 3D networks show great potential in facilitating this content creation process.

Immense progress has recently been seen in 3D-aware generative adversarial networks (GANs)~\cite{hologan,pigan,gram,eg3d,stylenerf,giraffe,xue2022giraffe,stylesdf}. However, articulated humans synthesized by 3D GANs still suffer from limited diversity and quality~\cite{grigorev2021stylepeople,bergman2022gnarf,yang20223dhumangan,sun2023next3d,noguchi2022unsupervised,jiang2023humangen,hong2023evad}. These limitations can be primarily attributed to either the limited representational capacity or computational inefficiency of existing 3D network architectures. For instance, some recent approaches~\cite{grigorev2021stylepeople,yang20223dhumangan,sun2023next3d} generate 2D features using application-specific template meshes, such as SMPL~\cite{loper2015smpl} for human bodies. These templates, unfortunately, cannot adequately model fine details like hair, clothes, or accessories. On the other hand, 3D GANs utilizing volumetric representations~\cite{bergman2022gnarf,noguchi2022unsupervised,jiang2023humangen,hong2023evad} have the potential to capture off-surface details, but the required volume rendering is often slow. This computational inefficiency can fundamentally limit the quality and diversity of the network since training a 3D GAN necessitates rendering tens of millions of images, which quickly becomes computationally infeasible. Consequently, upsampling networks have been widely adopted, but they often lead to significant degradation of generated shape quality and adversely impact multi-view consistency of the synthesized assets~\cite{stylenerf,giraffe,xue2022giraffe,eg3d}.

In this work, we aim to combine the advantages of efficient template meshes with the high representational capacity of volumetric scene representations. To this end, we introduce the concept of surface volumes, i.e., volumetric manifolds with non-zero thickness centered around the surface of a template mesh. A surface volume encapsulates off-surface details and volumetric structures like hair or accessories that exist close to a template but are not adequately modeled by an infinitesimally thin surface. Contrary to conventional volumetric representations, surface volumes do not waste capacity and resources in empty space. More importantly, we can approximate these volumetric manifolds using a set of layered isosurfaces, each represented as an appropriately deformed version of the original template mesh. These layers are textured with color and transparency, allowing for fast rasterization instead of slow volumetric ray casting to render an image. The RGBA textures of our layered surface volumes (LSVs) can be synthesized using conventional 2D generators. Remarkably, rendering an LSV is so efficient that the GAN training is now bottlenecked by texture generation rather than neural rendering, eliminating the need for view-inconsistent upsampling networks. LSVs draw inspiration from multiplane images~\cite{zhou2018stereo,single_view_mpi} and manifolds~\cite{gram,xiang2022gram}, but they are aligned with application-specific template meshes tailored to digital humans. By leveraging this novel representation in a 3D GAN setting, we demonstrate state-of-the-art quality, diversity, and multi-view consistency in generating articulated 3D humans on the \fashion~\cite{liu2016deepfasion} and StyleGAN-Human (\shhq)~\cite{fu2022stylegan} datasets.

\begin{figure}[t]
\includegraphics[width=\textwidth]{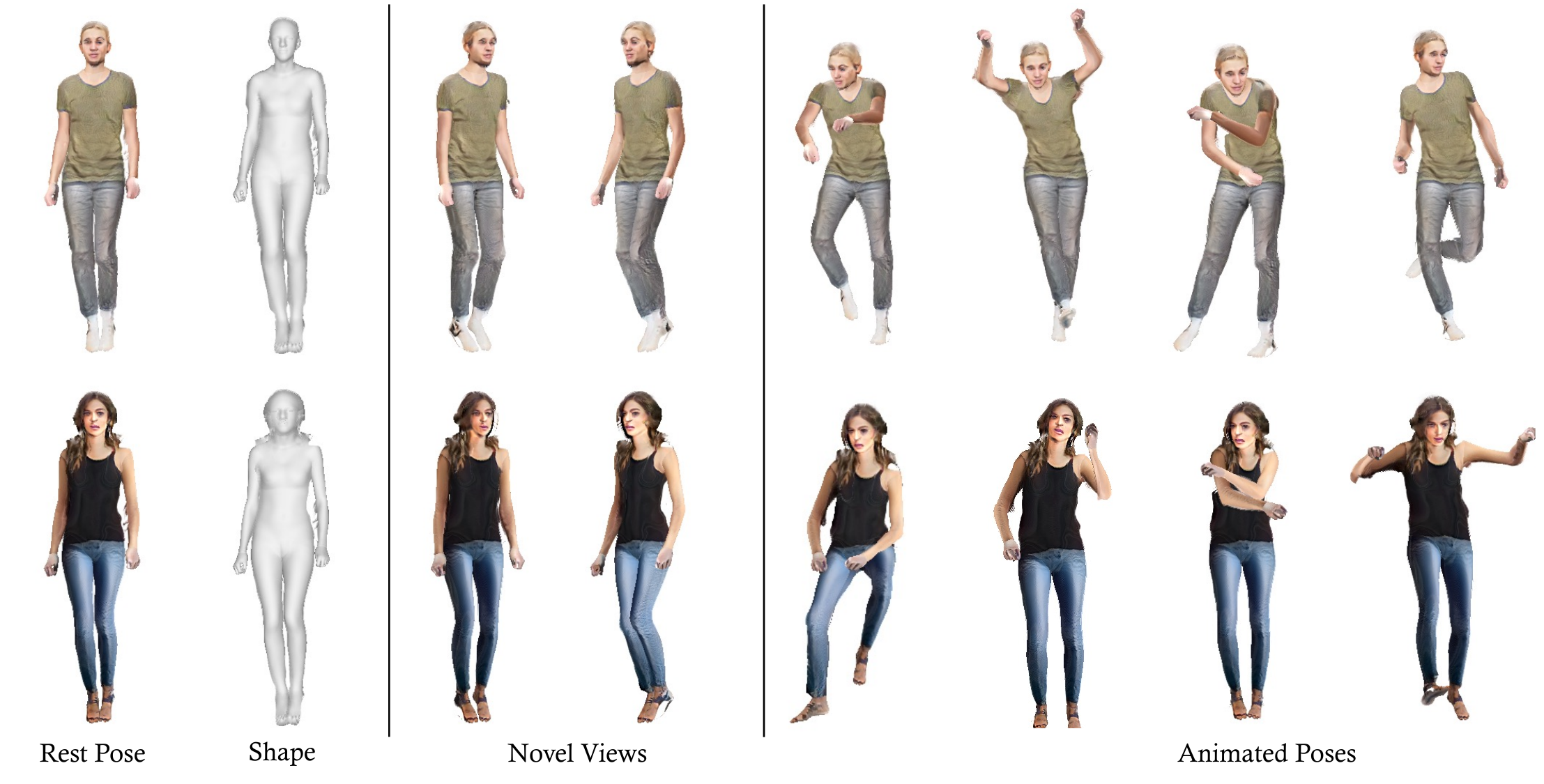} 
\caption{Trained using unstructured, single-view image collections, such as StyleGAN-Human~\cite{fu2022stylegan}, our GAN framework leverages a new layered surface volume representation to generate high-quality 3D human bodies in a canonical pose (left), which can be rendered from different camera perspectives (center), and animated using articulated motion (right). 
}
\label{teaser}
\vspace{-10pt}
\end{figure}

\section{Related Work}
\label{sec:related}

In this section, we briefly review the most relevant 
3D generative approaches. For a recent survey of the larger field of neural scene representation and rendering, we refer to~\cite{tewari2022advances}.

\paragraph{3D-aware Generative Models.}

Recent works on 3D GANs extend 2D image-based GANs~\cite{radford2015unsupervised,stylegan,stylegan2,stylegan3,fruhstuck2022insetgan,fu2022stylegan} by learning to generate 3D-aware multi-view-consistent objects or scenes from collections of unstructured, single-view 2D images in an unsupervised manner. The choice of neural scene representations has played a crucial role in the success of these 3D GANs. For example, some methods use meshes~\cite{Szabo:2019,Liao2020CVPR}, dense~\cite{wu2016learning,Gadelha:2017,VON,henzler2019platonicgan,hologan,nguyen2020blockgan,volumegan} or
sparse~\cite{hao2021GANcraft,schwarz2022voxgraf} voxel grids, 2D feature planes~\cite{eg3d,devries2021unconstrained,skorokhodov2022epigraf,bautista2022gaudi,son2022singraf,an2023panohead} or manifolds~\cite{gram,xiang2022gram,zhao2022generative}, fully implicit networks~\cite{graf,pigan,stylesdf,cips3d,shadegan,sun2021fenerf,d3d,shi20223d,zhang2022multi,chen2022sofgan,zhang20223d}, or a combination of low-resolution voxel grids combined with 2D CNN-based image upsampling layers~\cite{stylenerf,giraffe,xue2022giraffe,xu2022discoscene}. Very recently, diffusion models have also been explored as a platform for generating 3D objects or scenes~\cite{poole2022dreamfusion,wang2022score,lin2022magic3d,shue20223d}. 

Among these approaches, 3D GANs building on 2D feature planes or manifolds produce state-of-the-art multi-view-consistent image quality, approaching photorealism. Ours is most closely related to these methods, but rather than uniformly distributing the limited capacity of the 2D feature manifolds in 3D space, the proposed LSV representation focuses its capacity on a thin volume that is aligned with the surface of a template mesh for digital humans.

\paragraph{Generating Articulated 3D Digital Humans.}  

3D-aware GANs have been proposed to generate 3D digital humans whose body pose can be explicitly controlled by a user after generating the identity. 
Many recent approaches in this category generate a feature volume using a global~\cite{bergman2022gnarf,noguchi2022unsupervised,jiang2023humangen,Avatargen2023} or local~\cite{hong2023evad} triplane-based representation, which contains the human in a canonical pose and can be deformed using a target body pose. Another class of methods generates 2D textures or features on the surface of a human template mesh~\cite{grigorev2021stylepeople,yang20223dhumangan,sun2023next3d,aneja2022clipface}. Instead of generating only the appearance with a fixed template mesh, both shape and appearance can also be generated simultaneously~\cite{gao2022get3d}. Shape generation followed by text-guided texture optimization, for example using CLIP~\cite{hong2022avatarclip} or diffusion models~\cite{cao2023dreamavatar,zhang2023dreamface}, is also an emerging topic for digital human generation. Note that several of these works are concurrently developed to ours, without public code available (e.g.,~\cite{jiang2023humangen,Avatargen2023,jiang2023humangen,cao2023dreamavatar,zhang2023dreamface,aneja2022clipface}). 

Similar to many of these approaches, ours uses a template mesh for digital humans to enable post-generation articulation. Rather than using a single shell of such a template mesh, which limits its ability to represent hair, accessories, and other details, we introduce LSVs as a layered template mesh that combines the computational efficiency of mesh-based rendering with the flexibility of local volumes, allocated where needed, to represent fine detail. 
\section{Method}\label{sec:method}

Existing 3D GANs for articulated humans~\cite{bergman2022gnarf,noguchi2022unsupervised,jiang2023humangen,hong2023evad,Avatargen2023} often employ inefficient voxel representations and volume rendering, hindering performance during GAN training. In this section, we first introduce our layered surface volumes (LSVs) and the associated fast rasterization-based rendering pipeline as a way to alleviate these shortcomings. We then discuss how to use LSVs as a backbone in a 3D~GAN setting for generating digital humans. 

Code and pre-trained checkpoints will be made public.

\subsection{Layered Surface Volumes}
\label{sec:method:LSV}

\paragraph{Representation.}

We utilize a parametric mesh template to capture the generic shape of a human body, and leverage the pre-computed skinning weights and UV mapping to efficiently articulate and generate appearance of the human from texture maps.

In particular, we adopt SMPL~\cite{loper2015smpl}, which models the vertex locations of a human template mesh \(\mesh=\left( \verts, \faces \right)\) for body pose \(\pose\in\R^{3J}\) and shape \(\shape\in\R^{B}\), where \(\verts\in\R^{3V}\) and \(\faces\in\R^{3F}\) denote the vertex coordinates and face indices of the mesh, and \(J\) and \(B\) denote the number of joints and shape bases, respectively.
Formally, the SMPL model can be written as a mapping \(\verts = \smpl(\pose, \shape) \) with \(\smpl: \R^{|\pose|} \times \R^{|\shape|}\mapsto\R^{3V}\).
\(\smpl\) includes a shape-dependent articulation step based on linear blend skinning (LBS)~\cite{lewis2000pose} that deforms the mesh vertices from the T-pose to an arbitrary target pose:
\(\verts = \lbs\left( \mesh^{T}\left( \shape \right), \jointregressor\left( \shape \right), \pose \right)\), where \(\mesh^T\left( \shape \right)\) and \(\jointregressor\left( \shape \right)\) denote the T-pose mesh and the regressed joint locations, respectively.
Given the fixed topology and its parametrization, the appearance of a SMPL shape can be modeled through texture maps in the 2D UV space.
Each vertex $\v$ is assigned to a unique position in the 2D texture map \(\texture\), \ie,
\(\{\vcolor, o\} = \texture\left( \uv\left( \v \right) \right)\), where \(\vcolor\) and \(o\) are the retrieved color and opacity, \(\uv\) defines the mapping from vertex to coordinates on the texture map, which is pre-computed and fixed for the SMPL mesh.

\begin{figure}[t]
\includegraphics[width=\textwidth]{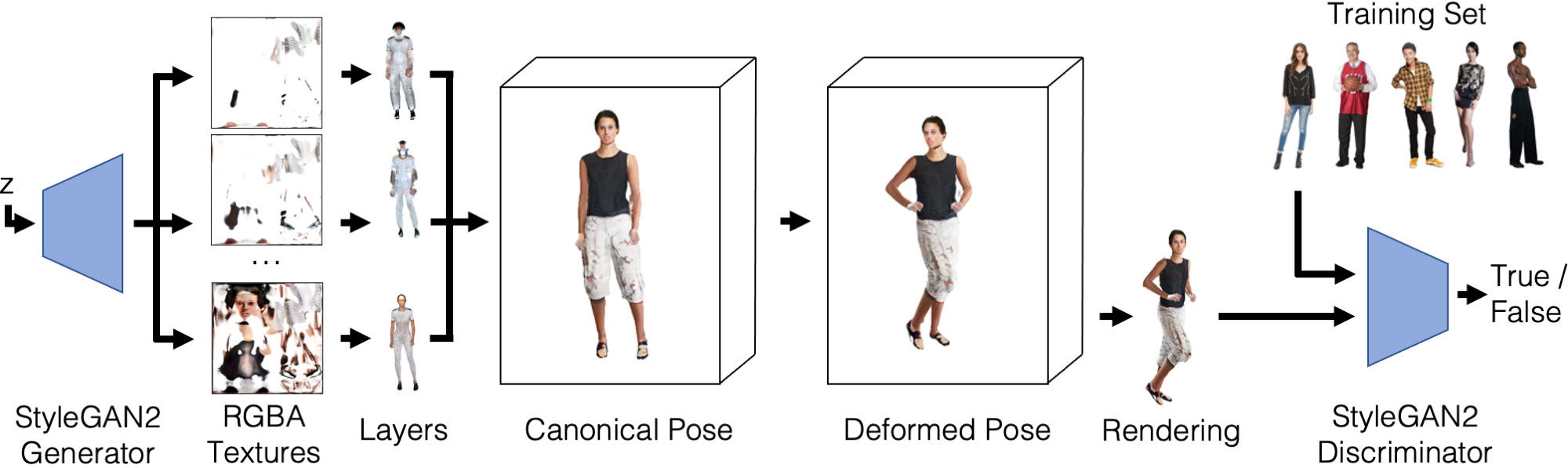} 
\caption{LSV-GAN pipeline. A latent code $z$ is fed into a 2D StyleGAN2 generator network, which outputs $N$ RGBA textures. These are applied to the individual mesh layers. All textured layers together are deformed into the target pose distribution and rendered using fast, differentiable rasterization before being fed into a camera- and body-pose-conditioned StyleGAN2 discriminator. An additional face discriminator is used but not shown.}
\label{fig:pipeline}
\end{figure}

The key idea of \moniker{} is to augment the base SMPL mesh \(\mesh^{T}\) with a small numbers of SMPL meshes \(\lbrace\mesh^T_{n}\rbrace_{n=1}^{N}\), namely \emph{layers}, wrapping around the base mesh.
Each layer is equipped with its own texture map \(\texture_{n}\in\R^{H\times W \times 4}\), encoding color \(\vcolor\in\R^{3}\) and opacity \(\density\in\R^{\geq0}\), which are composited together in the rendering stage (detailed in the next section) to capture geometry variations that can not be covered by a single SMPL base mesh.

Since the mesh topology stays the same, we can obtain the colors \(\vcolor_{n}\) and opacity \(\density_{n}\) of each layer from the texture map \(\texture_{n}\) using the same UV mapping. Let \(\texLayers\) denote all layers in LSVs, and \(\lbrack\colorSeq, \densitySeq\rbrack\) be color and opacity samples for vertex \(\v\) on all layers, 
\begin{equation}
\lbrack\colorSeq, \densitySeq\rbrack = \texLayers\left( \uv\left( \v \right) \right).\label{eq:texture_lookup}
\end{equation}
Similarly, the same skinning weights and joint regressor can be applied to simultaneously deform all mesh layers:
\begin{equation}
   \verts_{n} = \lbs \left( \mesh^{T}_{n}\left( \shape \right), \jointregressor\left( \shape \right), \pose \right).\label{eq:layer_deform}
\end{equation}

To obtain layers from a base mesh, we keep the connected topology in SMPL and inflate (and shrink) the mesh along the vertex normals \(\vn\) for a fixed thickness \(t_n\in\R\):
\begin{equation}
    \v_{n} = \v + t_n \mathbf{n}.\label{eq:layer_creation}
\end{equation}
We obtain \(\nMesh\) SMPL layers \(\lbrace\mesh_{n}\rbrace_{n=0}^{N-1}\), from the smallest to the largest, by setting $t_n = t_{min} + n (t_{max} - t_{min})/(N-1)$, with $t_{max}=0.01$ and $t_{min}=-0.01$.

\paragraph{Rendering.}

One key advantage of our representation lies in its rendering efficiency.
Instead of sampling the entire 3D volume hundreds of times along each ray, as done in volumetric rendering, we can apply differentiable rasterization~\cite{Laine2020diffrast}, which is much more efficient and has been highly optimized in hardware-accelerated graphics pipelines.
Specifically, to render an LSV model, we rasterize each layer independently.
For each pixel \(\pix\) on the final rendered image, the rasterizer finds the projected polygon (if any) with each layer, and evaluates the corresponding depth \(z_{n}\left( \pix \right)\), \(\vcolor_{n}\left( \pix \right)\) and opacity \(\density_{n}\left( \pix \right)\) by interpolating values the from the polygon vertices.
In what follows, we omit the pixel index \(\pix\) for the sake of readability.

We propose the following composition function to obtain the final color \(\vcolor\) at each pixel:
\begin{equation}
\vcolor = \sum_{n=1}^{N}\weight_{n}\density_{n}\vcolor_{n},\label{eq:pixel_color}
\end{equation}
where the compositing weights \(\weight_{n}\) depend on the relative depth of the layers
and their opacity, similar to \cite{lassner2021pulsar}:
\begin{equation}
w_{n} = \dfrac{\density_{n}\exp\left( -\density_{n}\bar{z}_{n}/\gamma \right)}{\sum_{n=1}^{N}\density_{n}\exp\left( -\density_{n}\bar{z}_{n}/\gamma \right)}, \textrm{ with }
\bar{z}_{n} = \frac{z_{n}-\min_{n}\left( z_{n} \right)}{\max_{n}\left( z_{n}\right)-\min_{n}\left( z_{n}\right)}\label{eq:composition}.
\end{equation}

\subsection{3D GAN Framework}
\label{sec:method:gan}

An overview of our generation framework can be found in \cref{fig:pipeline}. 
In this section, we elaborate on how to efficiently adopt LSVs into a 3D GAN framework for articulated humans.

\textbf{\noindent{Generator}}.
The textures of LSVs in our 3D GAN are generated by a StyleGAN2-based architecture,
%
which is tasked with generating color and opacity values for each layer, leading to a total of $4N$ output channels, instead of $3$.
The generator does not use camera pose or human pose conditioning, to prevent the rendered images from being overly dependent on the input view and body pose,
which helps to ensure that the generated textures are robust across different camera views and human poses when performing animation or camera movement.

\textbf{\noindent{Discriminator}}.
Our framework also leverages the discriminator $\mathrm{D(\cdot)}$ of StyleGAN2 for adversarial training.
Like EG3D~\cite{eg3d} and GNARF~\cite{bergman2022gnarf}, the discriminator is conditioned on camera and body poses, which enforces the synthesized images to be well-aligned with the given camera and pose condition instead of just being in the correct distribution.

\textbf{\noindent{Face Discriminator}}.
Given that the face occupies a small portion of the rendered image, the discriminator provides weak learning signal to the face region, which often times leads to inferior face quality.
We use the joints of SMPL to estimate a coarse bounding box of the face and crop the face patch from the whole frame. 
A face discriminator $\mathrm{D_{face}(\cdot)}$ is then introduced on the cropped face patch to help the texture generator learn better facial details.

\textbf{\noindent{Progressive Training}}.
High-fidelity texture maps are critical to the final rendering quality. However, synthesizing high-resolution textures poses a significant challenge to the generator. 
To address this issue, we adopt a progressive strategy for our GANs training. 
Unlike previous approaches that progressively add new blocks to both the generator and discriminator~\cite{karras2017progressive}, our method fixes the architecture of the texture generator while gradually increasing the resolution of the rendered image. 
This approach allows us to reuse the generator's parameters without requiring re-optimization when the resolution changes, resulting in more stable training. 
With our progressive training, the generator initially learns to synthesize coarse textures from low-resolution images, which then serve as good initialization for optimizing the textures into high-quality ones with fine-grained details. 
%

\textbf{\noindent{Hand Regularization}}.
Rendering realistic hands is another challenging task, particularly because the SMPL model cannot accurately simulate the distribution of real hands in the datasets (see supplement). 
Even if we deform the hand mesh, it cannot perfectly fit the actual hand pose and shape, resulting in artifacts such as translucent fingers. 
To overcome this issue, we reduce the deformation scale of the hand and process the hand textures $T_{\rm hand}$ independently using the UV atlas. 
To prevent the texture of the hand from learning to be transparent, we regularize the alpha map of the hand to be as opaque as possible with l$_1$ loss $\mathcal{L}_{\rm hand} = | \densitySeq_{\texLayers_{\rm hand}} - 1|$.
With this regularizer, we encourage hand textures to produce realistic and coherent results.

\textbf{\noindent{Training Details}}.
We first sample SMPL pose parameters $\mathbf{p}=(\beta, \gamma)$ from the dataset and generate the mesh layers according to our LSV representation. 
Next, we use a differentiable rasterizer~\cite{Laine2020diffrast} to render each textured isosurface into 2D images $\mathbf{I}_f$ and compose them as mentioned before.
The entire image generation process is formulated as $\mathbf{I}_f=\mathrm{G}(\mathbf{p}, \mathbf{z}, \xi)$, where the generator $\mathrm{G}(\cdot)$ takes as input SMPL parameters $\mathbf{p}$, a latent code $\mathbf{z}$ sampled from $\mathcal{N}(0, 1)$, and a camera pose $\xi$ to synthesize the image $\mathbf{I}_f$. 
During training, we randomly sample $\mathbf{p}$, $\mathbf{z}$, and $\xi$, while the real image $\mathbf{I}_r$ is sampled from the dataset.
Operating on the output of the generator, we use a discriminator $\mathrm{D}(\cdot)$ to guarantee the global coherence of the rendered human and a face discriminator $\mathrm{D_{face}}(\cdot)$ on cropped faces to improve face fidelity.
The generator and discriminators are jointly trained using $\mathcal{L}_{\rm GAN}$, the non-saturating GAN loss with R1 regularization on both discriminators.
Additionally, we use the aforementioned hand regularizer with weight $\lambda$ to improve the realism of hands.
The overall loss function is defined as $\mathcal{L}_{\rm total} = \mathcal{L}_{\rm GAN} + \lambda \mathcal{L}_{\rm hand}$.

\begin{figure}[t]
\centering
\includegraphics[width=0.95\textwidth]{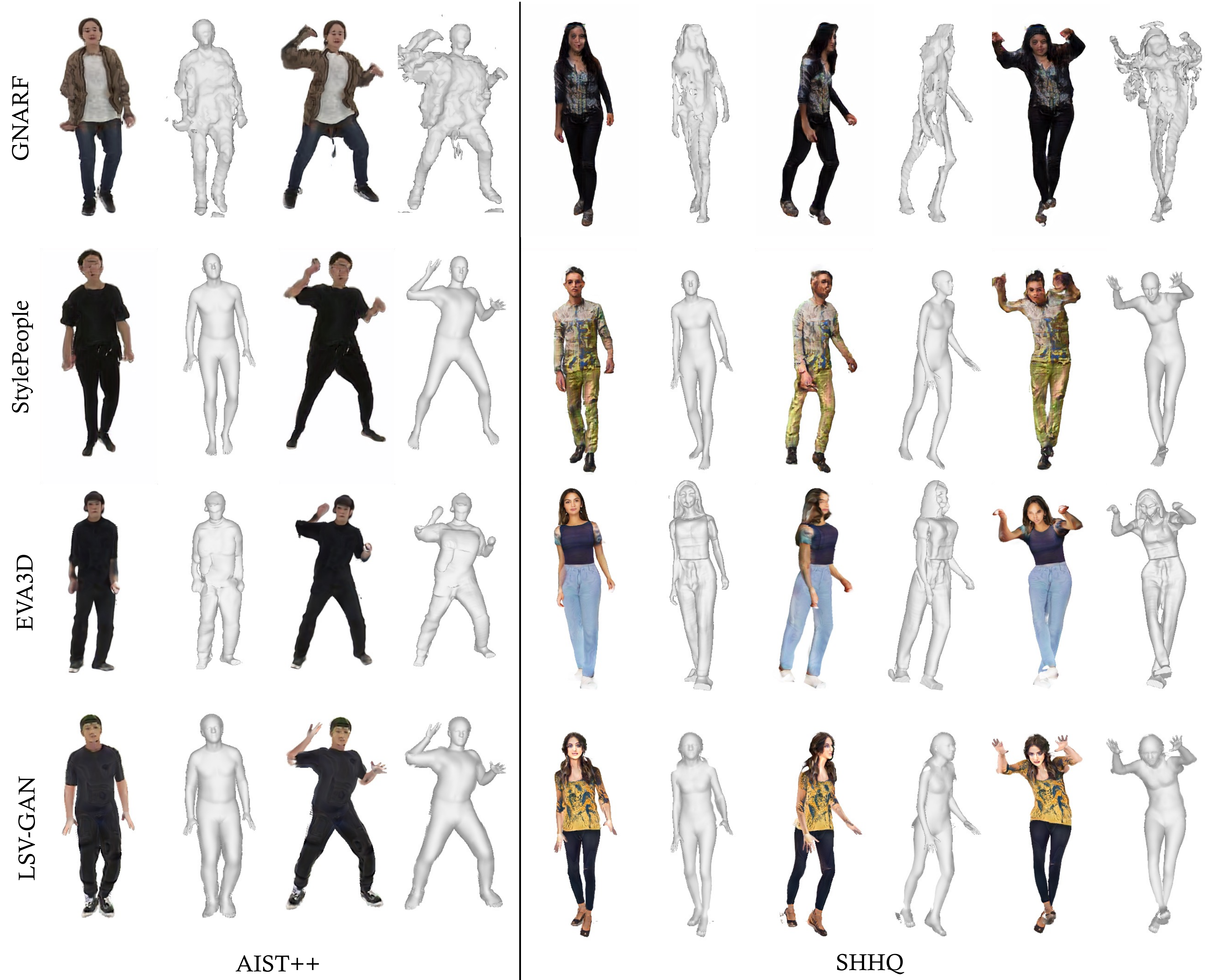}
\caption{Qualitative results and comparisons. We compare the results of several baselines, including GNARF, a representative implementation of StylePeople, and EVA3D, with our LSV-GAN using the \aist and \shhq datasets. In each case, we show an image and the shape rendered from the generated, canonical pose on the left in addition to one or two additional deformed body poses on the right. Our approach generates high-quality 3D humans with more detailed faces and more accurate shapes than the baselines.}
\label{fig:results}
\end{figure}

\section{Experiments}\label{sec:exp}

\subsection{Settings}

We first outline the settings of our experiments before evaluating the proposed LSV-based 3D GAN framework for articulated human generation. More implementation details can be found in the supplement.

\textbf{\noindent{Datasets.}}
We evaluate \method on three human datasets: \aist~\cite{li2021aist}, \fashion~\cite{liu2016deepfasion}, and \shhq~\cite{fu2022stylehuman}.
\aist is a large dataset consisting of 10.1M images covering 30 different performers in 9 camera views.
Each frame is annotated with a camera pose and fitted SMPL body poses.
We filter out the noisy samples with inaccurate SMPL annotations and collect a subset of 360$k$ images.
We also use the annotated bounding box to perform center cropping and then resize all images to a resolution of $512\times512$.
\fashion and \shhq are single-view image datasets consisting of 8k and 40k identities, respectively.
We adopt SMPLify-X~\cite{Pavlakos2019smplx} to estimate SMPL parameters and camera parameters.
All images from these two datasets are resized to $512\times256$ for GAN training.

\textbf{\noindent{Baselines.}}
We compare \method with several baselines:
EG3D~\cite{eg3d} and StyleSDF~\cite{stylesdf} are state-of-the-art methods for 3D-aware object synthesis;
ENARF~\cite{noguchi2022unsupervised} and GNARF~\cite{eg3d} are methods which perform deformation on triplane representation to achieve articulated human generation;
and EVA3D~\cite{hong2023evad} is a method which uses a compositional signed distance function for articulated human generation.
We also include a comparison with our inofficial implementation of StylePeople~\cite{grigorev2021stylepeople}, combining a single mesh layer and a neural feature decoder for human generation.
We implement StylePeople with our 1-layer surface volume rasterized at a resolution of $128 \times 128$ and then adopt a $4\times$ upsammpler~\cite{eg3d} to get the $512 \times 512$ output\footnote{The official repository of StylePeople only contains inference code. We implement and train it on new datasets for a fair comparison.}.    

\textbf{\noindent{Metrics.}}
We employ the Fréchet Inception Distance (FID) score to assess the quality and diversity of our generated images.
Specifically, we compute the FID score between 50,000 generated samples and all real images.
In addition, we use the Percentage of Correct Keypoints (PCK@0.5)~\cite{andriluka20142d} to evaluate the quality of animations and view-consistency of generated results. We use 5,000 samples to evaluate this, following the protocol in EVA3D.




\subsection{ Results}

\begin{table*}[t]
\center
\caption{Quantitative evaluation. We compare several baselines (left) using three different datasets. The quality and diversity, as measured by the FID score, are best for our LSV-GAN for the larger \fashion and \shhq datasets. Multi-view consistency is evaluated using the PCK metric; our approach consistently outperforms baselines in this metric. The training time (TR., measured in days on a single A6000 GPU) is the lowest for our method among all the high-resolution GANs operating at a resolution of $512^2$. The rendering time at inference (INF., measured in ms/image) is by far the lowest for our approach. $^*$~numbers adopted from~\cite{hong2023evad}; $^\dagger$ representative implementation of 2D texture generation on SMPL template mesh with feature-based upsampling, such as StylePeople. } \label{tab:main-results}
\vspace{-10pt}
\small
\begin{tabular}{lcccccccc}
\multicolumn{9}{c}{} \\
\toprule
\multirow{2}{*}{Model} & \multicolumn{2}{c}{\aist} & \multicolumn{2}{c}{\fashion} & \multicolumn{2}{c}{\shhq} & \multicolumn{2}{c}{Comp. Cost} \\
\cmidrule(lr){2-3}
\cmidrule(lr){4-5}
\cmidrule(lr){6-7}
\cmidrule(lr){8-9}
& FID $\downarrow$  &  PCK $\uparrow$  &  FID $\downarrow$   &  PCK $\uparrow$  & FID $\downarrow$   &  PCK $\uparrow$  & TR. $\downarrow$& INF. $\downarrow$\\
\midrule
EG3D$^*$ (512$^2$) & 34.76 & --- & 26.38 & --- & 32.96 & --- & 56 & 38\\
StyleSDF$^*$ (512$^2$) & 199.5 & --- & 92.40 & --- & 14.12 & --- & 65 & 32 \\ \midrule
ENRAF$^*$ (128$^2$)  & 73.07 & 42.85 & 77.03 & 43.74 & 80.54 &  40.17 &  {\bf 5} &  104  \\
GNARF (512$^2$)  & {\bf 11.13} & 96.11 & 33.85 & 97.83 & 14.84 & \underline{98.96} & 24 & \underline{72}\\
EVA3D$^*$ (512$^2$) & 19.40 & 83.15   & \underline{15.91} & 87.50 & \underline{11.99} & 88.95 & 40 & 200\\
StylePeople$^\dagger$ (1 layer, 512$^2$) & 18.97 & \underline{96.96} & 17.72 & \underline{98.31} &14.67  & 98.58 & \underline{20} & {\bf 28} \\
LSV-GAN (12 layers, 512$^2$)  & \underline{17.05} & {\bf 98.95} & {\bf 12.02} & {\bf 99.47}& {\bf 11.10} &  {\bf 99.44} & \underline{20} & {\bf 28} \\ 
\bottomrule
\end{tabular}
\vspace{-7pt}
\end{table*}

\paragraph{Qualitative Evaluation.}
\cref{fig:results} compares our method with all baselines at a resolution of 512 $\times$ 256.
We synthesize the images for all methods with the same SMPL pose for a fair comparison.
When using GNARF, which is trained at a low resolution and uses an image-based upsampler to achieve the target resolution, we observe inconsistency when synthesizing the same person with a variety of poses.
StylePeople also performs rasterization to map the neural textures onto a single-layer surface and then uses a neural decoder to generate human images.
This method achieves a limited quality and its 
decoder network, which models complex hair and accessories in 2D image space, introduces view-inconsistent artifacts.
While EVA3D is capable of generating high-quality images, the rendering of hands is not very detailed and we also witness blending artifacts when the arms are close to the body.
Additionally, rendering humans from a side view and performing large animations on EVA3D can result in artifacts on the face, hair, and hands, which is likely due to inaccurate geometry.
In comparison, our method generates 3D human images with high-fidelity appearance and holds better 3D consistency across different body poses.
Moreover, our model is able to handle challenging cases with large camera or body motion.
More detailed comparisons can be found in the supplementary material.
%
%

\paragraph{Quantitative Evaluation.}
As shown in \cref{tab:main-results}, \method consistently outperforms baselines in terms of all quantitative metrics.
EG3D and StyleSDF are not designed to handle the large diversity of body poses in the training data and humans generated with these approaches cannot be articulated.  
ENARF is trained at a low resolution, heavily relying on a view-inconsistent upsampling network, which results in low image quality.
In comparison to methods that adopt neural radiance fields (GNARF), our LSVs can directly generate high-resolution images without using any upsamplers or neural decoders, leading to superior image quality and multi-view consistency.
The large computational overhead of volumetric ray casting in EVA3D makes their training and inference cost much larger than other methods, despite good performance on image quality.
Moreover, the compositional representation used by EVA3D can lead to inconsistencies between the conditioned SMPL pose and the synthesized humans, as reflected by the PCK metric. 
In contrast, the PCK scores of our method are nearly 100\% across all datasets, demonstrating excellent multi-view consistency.
StylePeople also achieves high PCK scores, which are slightly impacted by the employed 2D feature decoder.
Additionally, our model can render full images without the need for an upsampling network, while maintaining better or comparable training and inference efficiency compared to other baselines operating at the same resolution. 
This is made possible by our layered surface volumes representation, which uses fast rasterization instead of slow volumetric rendering.
%
Note that the metric values of all models denoted with $^*$ in \cref{tab:main-results} are adopted from~\cite{hong2023evad}. Using the same evaluation procedure, we trained and evaluated the GNARF and StylePeople baselines as well as our method from scratch.

\begin{figure}[t]
\centering
\includegraphics[width=\textwidth]{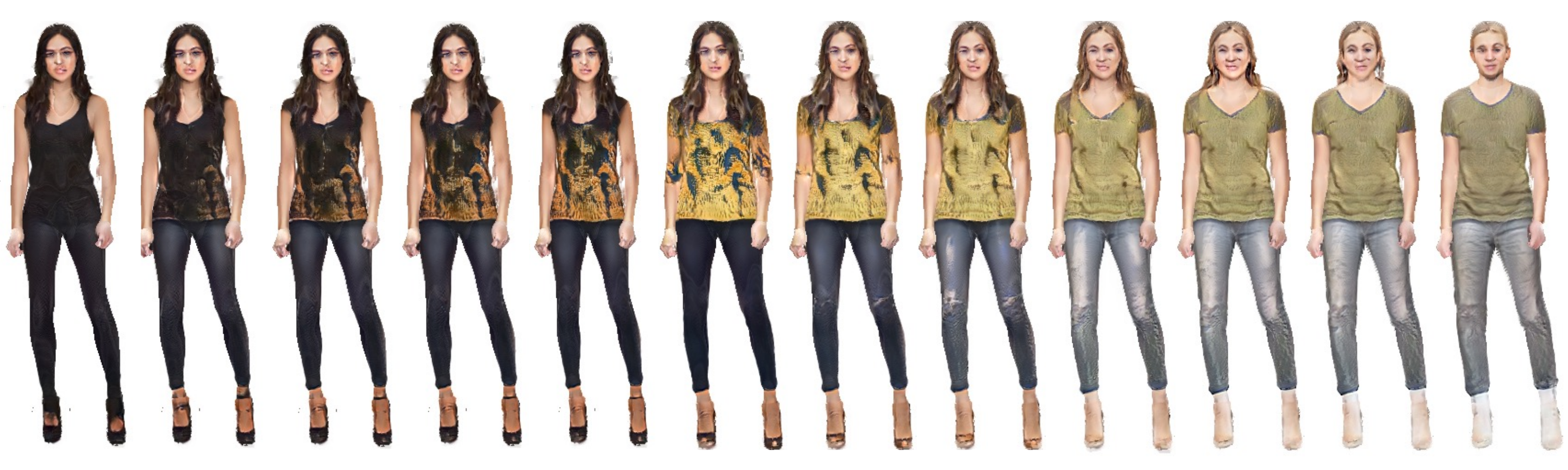}
\caption{Latent code interpolation of our approach trained on \shhq. }
\label{fig:interpolation}
\end{figure}

\paragraph{Latent Code Interpolation.}
In \cref{fig:interpolation}, we show example renderings of the interpolation between the latent codes of three different identities in the rest pose. This experiment validates the high quality of the latent space learned by LSV-GAN.

\begin{figure}[t]
    \centering
    \hspace{-3mm}
    \subfloat[Progressive training \label{fig:pgtrain}]{
        \includegraphics[width=0.33\textwidth]{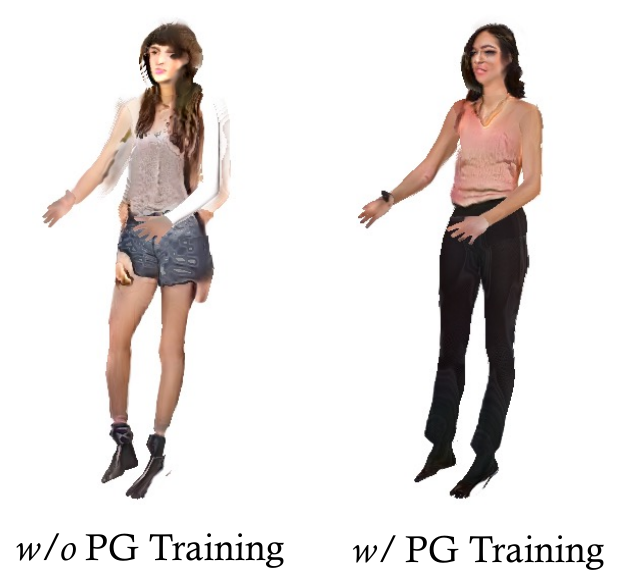}
    }\hspace{-2mm}
    \subfloat[Face discriminator \label{fig:facedist}]{
        \includegraphics[width=0.33\textwidth]{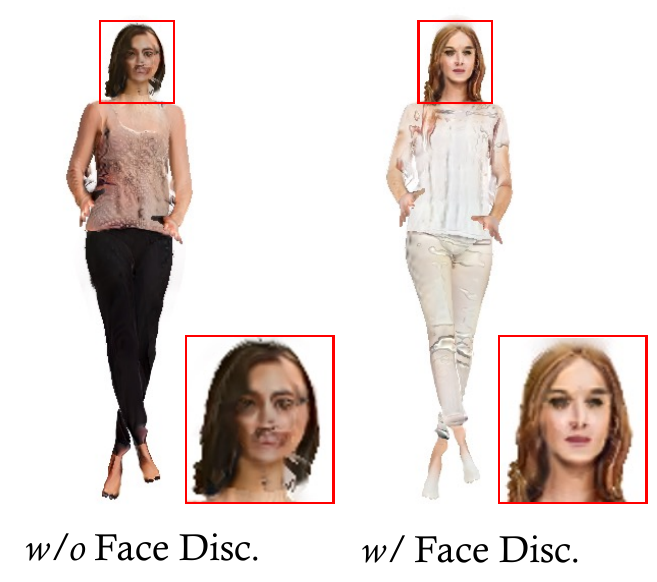}
    }\hspace{-2mm}
    \subfloat[Hand regularizer \label{fig:handreg}]{
        \includegraphics[width=0.33\textwidth]{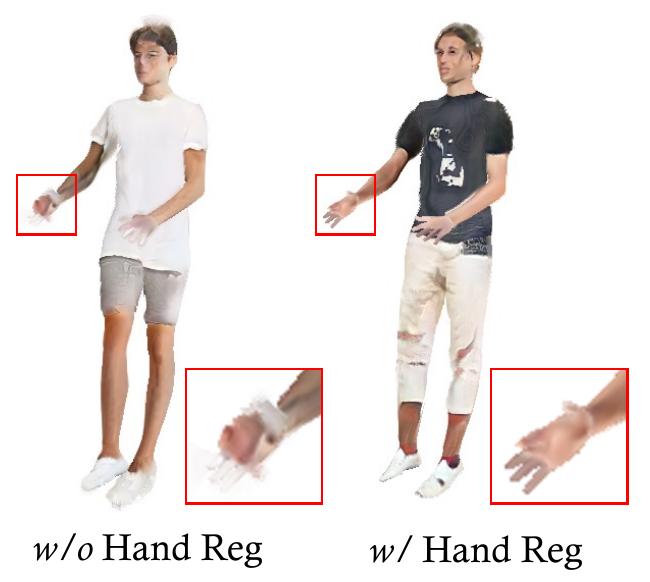}
    }
\caption{Qualitative comparison for ablations on progressive training (a), face discriminator (b), and hand regularizer (c).}
\label{fig:ablation}
\end{figure}

\subsection{Ablation Study}
We ablate the main components of \method to better understand their individual contributions.
Besides the FID score, we also include another metric, FID$_{\rm face}$, to evaluate the quality of generated faces.
All ablations are performed on \shhq with the same training schedule.

\paragraph{Number of Surface Layers.}
The number of layers of our surface volumes is a crucial factor in synthesizing realistic humans using our framework, as shown in \cref{tab:abalation-lsv}.
The single-layer case denotes a very basic setting where only the textured base SMPL layer is rasterized to render a human image.
However, the SMPL mesh does not account for deformed clothes, hair, and accessories, resulting in rendered images that deviate significantly from the real dataset distribution and produce poor FID and FID$_{\rm face}$ scores.
As we increase the number of layers, the quality and diversity of both images and faces improve significantly, as shown by the decreasing FID and FID$_{\rm face}$ scores.
As we show in the supplement, layered surface volumes can capture details and volumetric structures, such as hair and clothes, while a single-layer volume tends to generate very thin people and struggles to synthesize realistic human images.

\paragraph{Progressive Training.}
In \cref{fig:pgtrain}, we present a visual comparison between our model and the one trained without progressive training.
We observed that training the model at full resolution often leads to ``white texture'' artifacts, where the textures around the border of the arm or leg tend to learn the background color using opaque surface volumes to deceive the discriminator.
Progressive training, which starts at a low resolution, enables the optimization of a coarse texture initially.
This is beneficial for initializing a high-fidelity texture map later and also helps alleviate ``white texture'' artifacts.
\cref{tab:abalation-components} shows that progressive training also achieves better image quality than the base model, primarily due to the improvement of the texture quality.
\paragraph{Face Discriminator.}
In \cref{tab:abalation-components}, we present the effects of face discrimination.
We observe a significant improvement in face quality when face discrimination is applied.
The overall FID score also improves thanks to the good face fidelity.
The visual samples in \cref{fig:facedist} demonstrate that the model with face discriminator models hair as well as facial features with greater detail.

\paragraph{Hand Regularizer.}
We also ablate the hand regularizer to study its effects.
As shown in \cref{fig:handreg}, certain parts of the fingers are learned to be translucent to simulate complex hand poses without the hand regularization.
However, when utilizing a texture atlas to model hand textures independently and regularizing the alpha channel to be opaque, the rendered hands appear more natural.
Nevertheless, the quantitative results show a very slight drop in performance as the rendered hands do not fit the distribution of the real data as accurately, as shown in \cref{tab:abalation-components}. We nevertheless prefer using the hand regularizer, as it leads to more natural-looking results.

\begin{table}
\centering
\caption{Ablation study}
\captionsetup[subfloat]{captionskip=2pt}
\subfloat[Layers of Surface Volume \label{tab:abalation-lsv}]
{
\begin{tabular}{l|cc}
\toprule
\#LSV & FID & FID$_{\rm face}$ \\
\hline
1 & 101.3 & 124.3  \\
3 & 20.5 & 31.9  \\
6 & 15.8 &  27.7\\
12 & 11.1 & 24.6\\
\bottomrule
\end{tabular}
}
\hspace{2em}
\subfloat[Ablations on \method components \label{tab:abalation-components}]
{\begin{tabular}{l|cc}
\toprule
Model & FID & FID$_{\rm face}$  \\
\hline
N-layer base & 12.5 & 32.3\\
+ progressive training & 11.8 & 29.6 \\
+ face discriminator & 11.0 & 24.7\\
+ hand regularizer & 11.1 & 24.6\\
\bottomrule
\end{tabular}
}
\end{table}

\section{Discussion}
\label{sec:discussion}

\paragraph{Limitations and Future Work.}
Our work is limited in several ways. Although the quality, diversity, and view consistency of results generated with our approach are quantitatively better than the baselines, the level of detail for all 3D human GAN approaches is still relatively low. This limitation is primarily due to the limited resolution of $512^2$, which simply provides too few pixels for important body parts, such as faces. The resolution of 3D human GANs should be significantly improved, which could potentially be achieved using LSVs in combination with texture atlases or varying levels of detail in the generated textures. Accurate hand pose estimation from in-the-wild training images is challenging and inaccurate, which also degrades the quality and diversity of generated humans. Better pose estimation algorithms would help alleviate this issue. Although the textured layers generated by our method could in principle be directly imported into conventional graphics pipelines, we did not explore this direction. Finally, while the linear blend skinning approach used to animate our humans is fairly standard, it does not enable the realistic motion of hair, clothes, or other accessories. Combining LSVs with differentiable physical simulation engines could be an interesting avenue of future research. 

\paragraph{Ethical Considerations.}
GANs, such as ours, could be misused for generating edited imagery of real people. Such misuse of image
synthesis techniques poses a societal threat, and we do not condone using our work with the intent of spreading misinformation or tarnishing reputation. We also recognize a potential lack of diversity in
our results, stemming from implicit biases of the datasets we process.

\paragraph{Conclusion.}
We propose layered surface volumes (LSVs), a novel 3D representation for articulated digital humans, which combines the advantages of efficient template meshes with the high representational capacity of volumetric scene representations. Integrated with a 2D generator network architecture, our LSV-GAN overcomes the computational burden of neural volume rendering by leveraging fast rasterization, and is able to generate high-quality and view-consistent 3D articulated digital humans without the need for view-inconsistent 2D upsampling networks. These and other benefits of our framework enable us to take an important step towards generating photorealistic 3D digital human assets that can be articulated, which is a capability vital to the visual effects industry, virtual or augmented reality systems, and teleconferencing among other applications.

\vspace{5pt}
\noindent\textbf{Acknowledgements}. We thank Thabo Beeler, Sida Peng, Jianfeng Zhang, Fangzhou Hong, Ceyuan Yang for fruitful discussions and comments about this work.

{\small
\bibliographystyle{abbrvnat}
\bibliography{ref}

\begin{thebibliography}{71}
\providecommand{\natexlab}[1]{#1}
\providecommand{\url}[1]{\texttt{#1}}
\expandafter\ifx\csname urlstyle\endcsname\relax
  \providecommand{\doi}[1]{doi: #1}\else
  \providecommand{\doi}{doi: \begingroup \urlstyle{rm}\Url}\fi

\bibitem[An et~al.(2023)An, Xu, Shi, Song, Ogras, and Luo]{an2023panohead}
S.~An, H.~Xu, Y.~Shi, G.~Song, U.~Ogras, and L.~Luo.
\newblock Panohead: Geometry-aware 3d full-head synthesis in 360$^\circ$.
\newblock \emph{arXiv preprint arXiv:2303.13071}, 2023.

\bibitem[Andriluka et~al.(2014)Andriluka, Pishchulin, Gehler, and
  Schiele]{andriluka20142d}
M.~Andriluka, L.~Pishchulin, P.~Gehler, and B.~Schiele.
\newblock 2d human pose estimation: New benchmark and state of the art
  analysis.
\newblock In \emph{Proceedings of the IEEE Conference on computer Vision and
  Pattern Recognition}, pages 3686--3693, 2014.

\bibitem[Aneja et~al.(2022)Aneja, Thies, Dai, and Nießner]{aneja2022clipface}
S.~Aneja, J.~Thies, A.~Dai, and M.~Nießner.
\newblock {C}lip{F}ace: {T}ext-guided {E}diting of {T}extured 3{D} {M}orphable
  {M}odels.
\newblock In \emph{ArXiv preprint arXiv:2212.01406}, 2022.

\bibitem[Bautista et~al.(2022)Bautista, Guo, Abnar, Talbott, Toshev, Chen,
  Dinh, Zhai, Goh, Ulbricht, et~al.]{bautista2022gaudi}
M.~A. Bautista, P.~Guo, S.~Abnar, W.~Talbott, A.~Toshev, Z.~Chen, L.~Dinh,
  S.~Zhai, H.~Goh, D.~Ulbricht, et~al.
\newblock Gaudi: A neural architect for immersive 3d scene generation.
\newblock \emph{Advances in Neural Information Processing Systems},
  35:\penalty0 25102--25116, 2022.

\bibitem[Bergman et~al.(2022)Bergman, Kellnhofer, Yifan, Chan, Lindell, and
  Wetzstein]{bergman2022gnarf}
A.~W. Bergman, P.~Kellnhofer, W.~Yifan, E.~R. Chan, D.~B. Lindell, and
  G.~Wetzstein.
\newblock Generative neural articulated radiance fields.
\newblock In \emph{NeurIPS}, 2022.

\bibitem[Cao et~al.(2023)Cao, Cao, Han, Shan, and Wong]{cao2023dreamavatar}
Y.~Cao, Y.-P. Cao, K.~Han, Y.~Shan, and K.-Y.~K. Wong.
\newblock Dreamavatar: Text-and-shape guided 3d human avatar generation via
  diffusion models, 2023.

\bibitem[Chan et~al.(2021)Chan, Monteiro, Kellnhofer, Wu, and Wetzstein]{pigan}
E.~Chan, M.~Monteiro, P.~Kellnhofer, J.~Wu, and G.~Wetzstein.
\newblock pi-{GAN}: Periodic implicit generative adversarial networks for
  3d-aware image synthesis.
\newblock In \emph{CVPR}, 2021.

\bibitem[Chan et~al.(2022)Chan, Lin, Chan, Nagano, Pan, De~Mello, Gallo,
  Guibas, Tremblay, Khamis, et~al.]{eg3d}
E.~R. Chan, C.~Z. Lin, M.~A. Chan, K.~Nagano, B.~Pan, S.~De~Mello, O.~Gallo,
  L.~Guibas, J.~Tremblay, S.~Khamis, et~al.
\newblock Efficient geometry-aware 3d generative adversarial networks.
\newblock In \emph{CVPR}, 2022.

\bibitem[Chen et~al.(2022)Chen, Liu, Xie, Chen, Su, and Yu]{chen2022sofgan}
A.~Chen, R.~Liu, L.~Xie, Z.~Chen, H.~Su, and J.~Yu.
\newblock Sofgan: A portrait image generator with dynamic styling.
\newblock \emph{ACM Transactions on Graphics (TOG)}, 41\penalty0 (1):\penalty0
  1--26, 2022.

\bibitem[Deng et~al.(2022)Deng, Yang, Xiang, and Tong]{gram}
Y.~Deng, J.~Yang, J.~Xiang, and X.~Tong.
\newblock Gram: Generative radiance manifolds for 3d-aware image generation.
\newblock In \emph{CVPR}, 2022.

\bibitem[DeVries et~al.(2021)DeVries, Bautista, Srivastava, Taylor, and
  Susskind]{devries2021unconstrained}
T.~DeVries, M.~A. Bautista, N.~Srivastava, G.~W. Taylor, and J.~M. Susskind.
\newblock Unconstrained scene generation with locally conditioned radiance
  fields.
\newblock In \emph{CVPR}, pages 14304--14313, 2021.

\bibitem[Fr{\"u}hst{\"u}ck et~al.(2022)Fr{\"u}hst{\"u}ck, Singh, Shechtman,
  Mitra, Wonka, and Lu]{fruhstuck2022insetgan}
A.~Fr{\"u}hst{\"u}ck, K.~K. Singh, E.~Shechtman, N.~J. Mitra, P.~Wonka, and
  J.~Lu.
\newblock Insetgan for full-body image generation.
\newblock In \emph{CVPR}, pages 7723--7732, 2022.

\bibitem[Fu et~al.(2022{\natexlab{a}})Fu, Li, Jiang, Lin, Qian, Loy, Wu, and
  Liu]{fu2022stylegan}
J.~Fu, S.~Li, Y.~Jiang, K.-Y. Lin, C.~Qian, C.~C. Loy, W.~Wu, and Z.~Liu.
\newblock Stylegan-human: A data-centric odyssey of human generation.
\newblock In \emph{ECCV}, pages 1--19, 2022{\natexlab{a}}.

\bibitem[Fu et~al.(2022{\natexlab{b}})Fu, Li, Jiang, Lin, Qian, Loy, Wu, and
  Liu]{fu2022stylehuman}
J.~Fu, S.~Li, Y.~Jiang, K.-Y. Lin, C.~Qian, C.~C. Loy, W.~Wu, and Z.~Liu.
\newblock Stylegan-human: A data-centric odyssey of human generation.
\newblock In \emph{ECCV}, 2022{\natexlab{b}}.

\bibitem[Gadelha et~al.(2017)Gadelha, Maji, and Wang]{Gadelha:2017}
M.~Gadelha, S.~Maji, and R.~Wang.
\newblock {3D} shape induction from {2D} views of multiple objects.
\newblock In \emph{3DV}, 2017.

\bibitem[Gao et~al.(2022)Gao, Shen, Wang, Chen, Yin, Li, Litany, Gojcic, and
  Fidler]{gao2022get3d}
J.~Gao, T.~Shen, Z.~Wang, W.~Chen, K.~Yin, D.~Li, O.~Litany, Z.~Gojcic, and
  S.~Fidler.
\newblock Get3d: A generative model of high quality 3d textured shapes learned
  from images.
\newblock In \emph{NeurIPS}, 2022.

\bibitem[Grigorev et~al.(2021)Grigorev, Iskakov, Ianina, Bashirov, Zakharkin,
  Vakhitov, and Lempitsky]{grigorev2021stylepeople}
A.~Grigorev, K.~Iskakov, A.~Ianina, R.~Bashirov, I.~Zakharkin, A.~Vakhitov, and
  V.~Lempitsky.
\newblock Stylepeople: A generative model of fullbody human avatars.
\newblock In \emph{CVPR}, pages 5151--5160, 2021.

\bibitem[Gu et~al.(2022)Gu, Liu, Wang, and Theobalt]{stylenerf}
J.~Gu, L.~Liu, P.~Wang, and C.~Theobalt.
\newblock Stylenerf: A style-based 3d-aware generator for high-resolution image
  synthesis.
\newblock In \emph{Int. Conf. Learn. Represent.}, 2022.

\bibitem[Hao et~al.(2021)Hao, Mallya, Belongie, and Liu]{hao2021GANcraft}
Z.~Hao, A.~Mallya, S.~Belongie, and M.-Y. Liu.
\newblock {GANcraft}: {U}nsupervised {3D} neural rendering of minecraft worlds.
\newblock In \emph{ICCV}, 2021.

\bibitem[Henzler et~al.(2019)Henzler, Mitra, and
  Ritschel]{henzler2019platonicgan}
P.~Henzler, N.~J. Mitra, and T.~Ritschel.
\newblock Escaping {P}lato's cave: {3D} shape from adversarial rendering.
\newblock In \emph{ICCV}, 2019.

\bibitem[Hong et~al.(2022)Hong, Zhang, Pan, Cai, Yang, and
  Liu]{hong2022avatarclip}
F.~Hong, M.~Zhang, L.~Pan, Z.~Cai, L.~Yang, and Z.~Liu.
\newblock Avatarclip: Zero-shot text-driven generation and animation of 3d
  avatars.
\newblock \emph{ACM Transactions on Graphics (TOG)}, 41\penalty0 (4):\penalty0
  1--19, 2022.

\bibitem[Hong et~al.(2023)Hong, Chen, LAN, Pan, and Liu]{hong2023evad}
F.~Hong, Z.~Chen, Y.~LAN, L.~Pan, and Z.~Liu.
\newblock {EVA}3d: Compositional 3d human generation from 2d image collections.
\newblock In \emph{ICLR}, 2023.

\bibitem[Jiang et~al.(2023)Jiang, Jiang, Wang, Luo, Chen, and
  Xu]{jiang2023humangen}
S.~Jiang, H.~Jiang, Z.~Wang, H.~Luo, W.~Chen, and L.~Xu.
\newblock Humangen: Generating human radiance fields with explicit priors.
\newblock In \emph{CVPR}, 2023.

\bibitem[Karras et~al.(2017)Karras, Aila, Laine, and
  Lehtinen]{karras2017progressive}
T.~Karras, T.~Aila, S.~Laine, and J.~Lehtinen.
\newblock Progressive growing of gans for improved quality, stability, and
  variation.
\newblock \emph{arXiv preprint arXiv:1710.10196}, 2017.

\bibitem[Karras et~al.(2019)Karras, Laine, and Aila]{stylegan}
T.~Karras, S.~Laine, and T.~Aila.
\newblock A style-based generator architecture for generative adversarial
  networks.
\newblock In \emph{CVPR}, 2019.

\bibitem[Karras et~al.(2020)Karras, Laine, Aittala, Hellsten, Lehtinen, and
  Aila]{stylegan2}
T.~Karras, S.~Laine, M.~Aittala, J.~Hellsten, J.~Lehtinen, and T.~Aila.
\newblock Analyzing and improving the image quality of {StyleGAN}.
\newblock In \emph{CVPR}, 2020.

\bibitem[Karras et~al.(2021)Karras, Aittala, Laine, H\"ark\"onen, Hellsten,
  Lehtinen, and Aila]{stylegan3}
T.~Karras, M.~Aittala, S.~Laine, E.~H\"ark\"onen, J.~Hellsten, J.~Lehtinen, and
  T.~Aila.
\newblock Alias-free generative adversarial networks.
\newblock In \emph{NeurIPS}, 2021.

\bibitem[Laine et~al.(2020)Laine, Hellsten, Karras, Seol, Lehtinen, and
  Aila]{Laine2020diffrast}
S.~Laine, J.~Hellsten, T.~Karras, Y.~Seol, J.~Lehtinen, and T.~Aila.
\newblock Modular primitives for high-performance differentiable rendering.
\newblock \emph{ACM Transactions on Graphics}, 39\penalty0 (6), 2020.

\bibitem[Lassner and Zollhofer(2021)]{lassner2021pulsar}
C.~Lassner and M.~Zollhofer.
\newblock Pulsar: Efficient sphere-based neural rendering.
\newblock In \emph{Proceedings of the IEEE/CVF Conference on Computer Vision
  and Pattern Recognition}, pages 1440--1449, 2021.

\bibitem[Lewis et~al.(2000)Lewis, Cordner, and Fong]{lewis2000pose}
J.~P. Lewis, M.~Cordner, and N.~Fong.
\newblock Pose space deformation: a unified approach to shape interpolation and
  skeleton-driven deformation.
\newblock In \emph{Proceedings of the 27th annual conference on Computer
  graphics and interactive techniques}, pages 165--172, 2000.

\bibitem[Li et~al.(2021)Li, Yang, Ross, and Kanazawa]{li2021aist}
R.~Li, S.~Yang, D.~A. Ross, and A.~Kanazawa.
\newblock Learn to dance with aist++: Music conditioned 3d dance generation,
  2021.

\bibitem[Liao et~al.(2020)Liao, Schwarz, Mescheder, and Geiger]{Liao2020CVPR}
Y.~Liao, K.~Schwarz, L.~Mescheder, and A.~Geiger.
\newblock Towards unsupervised learning of generative models for {3D}
  controllable image synthesis.
\newblock In \emph{CVPR}, 2020.

\bibitem[Lin et~al.(2022)Lin, Gao, Tang, Takikawa, Zeng, Huang, Kreis, Fidler,
  Liu, and Lin]{lin2022magic3d}
C.-H. Lin, J.~Gao, L.~Tang, T.~Takikawa, X.~Zeng, X.~Huang, K.~Kreis,
  S.~Fidler, M.-Y. Liu, and T.-Y. Lin.
\newblock Magic3d: High-resolution text-to-3d content creation.
\newblock \emph{arXiv preprint arXiv:2211.10440}, 2022.

\bibitem[Liu et~al.(2016)Liu, Luo, Qiu, Wang, and Tang]{liu2016deepfasion}
Z.~Liu, P.~Luo, S.~Qiu, X.~Wang, and X.~Tang.
\newblock Deepfashion: Powering robust clothes recognition and retrieval with
  rich annotations.
\newblock In \emph{CVPR}, 2016.

\bibitem[Loper et~al.(2015)Loper, Mahmood, Romero, Pons-Moll, and
  Black]{loper2015smpl}
M.~Loper, N.~Mahmood, J.~Romero, G.~Pons-Moll, and M.~J. Black.
\newblock Smpl: A skinned multi-person linear model.
\newblock \emph{ACM transactions on graphics (TOG)}, 34\penalty0 (6):\penalty0
  1--16, 2015.

\bibitem[Nguyen-Phuoc et~al.(2019)Nguyen-Phuoc, Li, Theis, Richardt, and
  Yang]{hologan}
T.~Nguyen-Phuoc, C.~Li, L.~Theis, C.~Richardt, and Y.-L. Yang.
\newblock {HoloGAN}: Unsupervised learning of 3d representations from natural
  images.
\newblock In \emph{ICCV}, 2019.

\bibitem[Nguyen-Phuoc et~al.(2020)Nguyen-Phuoc, Richardt, Mai, Yang, and
  Mitra]{nguyen2020blockgan}
T.~Nguyen-Phuoc, C.~Richardt, L.~Mai, Y.-L. Yang, and N.~Mitra.
\newblock {BlockGAN}: Learning {3D} object-aware scene representations from
  unlabelled images.
\newblock In \emph{NeurIPS}, 2020.

\bibitem[Niemeyer and Geiger(2021)]{giraffe}
M.~Niemeyer and A.~Geiger.
\newblock {GIRAFFE}: Representing scenes as compositional generative neural
  feature fields.
\newblock In \emph{CVPR}, 2021.

\bibitem[Noguchi et~al.(2022)Noguchi, Sun, Lin, and
  Harada]{noguchi2022unsupervised}
A.~Noguchi, X.~Sun, S.~Lin, and T.~Harada.
\newblock Unsupervised learning of efficient geometry-aware neural articulated
  representations.
\newblock In \emph{ECCV}, pages 597--614, 2022.

\bibitem[Or-El et~al.(2022)Or-El, Luo, Shan, Shechtman, Park, and
  Kemelmacher-Shlizerman]{stylesdf}
R.~Or-El, X.~Luo, M.~Shan, E.~Shechtman, J.~J. Park, and
  I.~Kemelmacher-Shlizerman.
\newblock Stylesdf: High-resolution 3d-consistent image and geometry
  generation.
\newblock In \emph{CVPR}, 2022.

\bibitem[Pan et~al.(2021)Pan, Xu, Loy, Theobalt, and Dai]{shadegan}
X.~Pan, X.~Xu, C.~C. Loy, C.~Theobalt, and B.~Dai.
\newblock A shading-guided generative implicit model for shape-accurate
  3d-aware image synthesis.
\newblock In \emph{NeurIPS}, 2021.

\bibitem[Pavlakos et~al.(2019)Pavlakos, Choutas, Ghorbani, Bolkart, Osman,
  Tzionas, and Black]{Pavlakos2019smplx}
G.~Pavlakos, V.~Choutas, N.~Ghorbani, T.~Bolkart, A.~A.~A. Osman, D.~Tzionas,
  and M.~J. Black.
\newblock Expressive body capture: 3d hands, face, and body from a single
  image.
\newblock In \emph{CVPR}, 2019.

\bibitem[Poole et~al.(2022)Poole, Jain, Barron, and
  Mildenhall]{poole2022dreamfusion}
B.~Poole, A.~Jain, J.~T. Barron, and B.~Mildenhall.
\newblock Dreamfusion: Text-to-3d using 2d diffusion.
\newblock \emph{arXiv preprint arXiv:2209.14988}, 2022.

\bibitem[Radford et~al.(2015)Radford, Metz, and
  Chintala]{radford2015unsupervised}
A.~Radford, L.~Metz, and S.~Chintala.
\newblock Unsupervised representation learning with deep convolutional
  generative adversarial networks.
\newblock \emph{arXiv preprint arXiv:1511.06434}, 2015.

\bibitem[Schwarz et~al.(2020)Schwarz, Liao, Niemeyer, and Geiger]{graf}
K.~Schwarz, Y.~Liao, M.~Niemeyer, and A.~Geiger.
\newblock {GRAF}: Generative radiance fields for 3d-aware image synthesis.
\newblock In \emph{NeurIPS}, 2020.

\bibitem[Schwarz et~al.(2022)Schwarz, Sauer, Niemeyer, Liao, and
  Geiger]{schwarz2022voxgraf}
K.~Schwarz, A.~Sauer, M.~Niemeyer, Y.~Liao, and A.~Geiger.
\newblock Voxgraf: Fast 3d-aware image synthesis with sparse voxel grids.
\newblock \emph{arXiv preprint arXiv:2206.07695}, 2022.

\bibitem[Shi et~al.(2022)Shi, Shen, Zhu, Yeung, and Chen]{shi20223d}
Z.~Shi, Y.~Shen, J.~Zhu, D.-Y. Yeung, and Q.~Chen.
\newblock 3d-aware indoor scene synthesis with depth priors.
\newblock In \emph{ECCV}, pages 406--422. Springer, 2022.

\bibitem[Shue et~al.(2023)Shue, Chan, Po, Ankner, Wu, and
  Wetzstein]{shue20223d}
J.~R. Shue, E.~R. Chan, R.~Po, Z.~Ankner, J.~Wu, and G.~Wetzstein.
\newblock 3d neural field generation using triplane diffusion.
\newblock In \emph{CVPR}, 2023.

\bibitem[Skorokhodov et~al.(2022)Skorokhodov, Tulyakov, Wang, and
  Wonka]{skorokhodov2022epigraf}
I.~Skorokhodov, S.~Tulyakov, Y.~Wang, and P.~Wonka.
\newblock Epigraf: Rethinking training of 3d gans.
\newblock \emph{arXiv preprint arXiv:2206.10535}, 2022.

\bibitem[Son et~al.(2022)Son, Park, Guibas, and Wetzstein]{son2022singraf}
M.~Son, J.~J. Park, L.~Guibas, and G.~Wetzstein.
\newblock Singraf: Learning a 3d generative radiance field for a single scene.
\newblock \emph{arXiv preprint arXiv:2211.17260}, 2022.

\bibitem[Sun et~al.(2022)Sun, Wang, Zhang, Li, Zhang, Liu, and
  Wang]{sun2021fenerf}
J.~Sun, X.~Wang, Y.~Zhang, X.~Li, Q.~Zhang, Y.~Liu, and J.~Wang.
\newblock Fenerf: Face editing in neural radiance fields.
\newblock In \emph{CVPR}, 2022.

\bibitem[Sun et~al.(2023)Sun, Wang, Wang, Li, Zhang, Zhang, and
  Liu]{sun2023next3d}
J.~Sun, X.~Wang, L.~Wang, X.~Li, Y.~Zhang, H.~Zhang, and Y.~Liu.
\newblock Next3d: Generative neural texture rasterization for 3d-aware head
  avatars.
\newblock In \emph{CVPR}, 2023.

\bibitem[Szab\'o et~al.(2019)Szab\'o, Meishvili, and Favaro]{Szabo:2019}
A.~Szab\'o, G.~Meishvili, and P.~Favaro.
\newblock Unsupervised generative {3D} shape learning from natural images.
\newblock \emph{arXiv preprint arXiv:1910.00287}, 2019.

\bibitem[Tewari et~al.(2022{\natexlab{a}})Tewari, Pan, Fried, Agrawala,
  Theobalt, et~al.]{d3d}
A.~Tewari, X.~Pan, O.~Fried, M.~Agrawala, C.~Theobalt, et~al.
\newblock Disentangled3d: Learning a 3d generative model with disentangled
  geometry and appearance from monocular images.
\newblock In \emph{CVPR}, 2022{\natexlab{a}}.

\bibitem[Tewari et~al.(2022{\natexlab{b}})Tewari, Thies, Mildenhall,
  Srinivasan, Tretschk, Yifan, Lassner, Sitzmann, Martin-Brualla, Lombardi,
  et~al.]{tewari2022advances}
A.~Tewari, J.~Thies, B.~Mildenhall, P.~Srinivasan, E.~Tretschk, W.~Yifan,
  C.~Lassner, V.~Sitzmann, R.~Martin-Brualla, S.~Lombardi, et~al.
\newblock Advances in neural rendering.
\newblock In \emph{Computer Graphics Forum}, volume~41, pages 703--735. Wiley
  Online Library, 2022{\natexlab{b}}.

\bibitem[Tucker and Snavely(2020)]{single_view_mpi}
R.~Tucker and N.~Snavely.
\newblock Single-view view synthesis with multiplane images.
\newblock In \emph{CVPR}, 2020.

\bibitem[Wang et~al.(2022)Wang, Du, Li, Yeh, and Shakhnarovich]{wang2022score}
H.~Wang, X.~Du, J.~Li, R.~A. Yeh, and G.~Shakhnarovich.
\newblock Score jacobian chaining: Lifting pretrained 2d diffusion models for
  3d generation.
\newblock \emph{arXiv preprint arXiv:2212.00774}, 2022.

\bibitem[Wu et~al.(2016)Wu, Zhang, Xue, Freeman, and Tenenbaum]{wu2016learning}
J.~Wu, C.~Zhang, T.~Xue, B.~Freeman, and J.~Tenenbaum.
\newblock Learning a probabilistic latent space of object shapes via 3d
  generative-adversarial modeling.
\newblock \emph{Advances in neural information processing systems}, 29, 2016.

\bibitem[Xiang et~al.(2022)Xiang, Yang, Deng, and Tong]{xiang2022gram}
J.~Xiang, J.~Yang, Y.~Deng, and X.~Tong.
\newblock Gram-hd: 3d-consistent image generation at high resolution with
  generative radiance manifolds.
\newblock \emph{arXiv preprint arXiv:2206.07255}, 2022.

\bibitem[Xu et~al.(2022{\natexlab{a}})Xu, Chai, Shi, Peng, Skorokhodov,
  Siarohin, Yang, Shen, Lee, Zhou, et~al.]{xu2022discoscene}
Y.~Xu, M.~Chai, Z.~Shi, S.~Peng, I.~Skorokhodov, A.~Siarohin, C.~Yang, Y.~Shen,
  H.-Y. Lee, B.~Zhou, et~al.
\newblock Discoscene: Spatially disentangled generative radiance fields for
  controllable 3d-aware scene synthesis.
\newblock \emph{arXiv preprint arXiv:2212.11984}, 2022{\natexlab{a}}.

\bibitem[Xu et~al.(2022{\natexlab{b}})Xu, Peng, Yang, Shen, and
  Zhou]{volumegan}
Y.~Xu, S.~Peng, C.~Yang, Y.~Shen, and B.~Zhou.
\newblock 3d-aware image synthesis via learning structural and textural
  representations.
\newblock In \emph{CVPR}, 2022{\natexlab{b}}.

\bibitem[Xue et~al.(2022)Xue, Li, Singh, and Lee]{xue2022giraffe}
Y.~Xue, Y.~Li, K.~K. Singh, and Y.~J. Lee.
\newblock Giraffe hd: A high-resolution 3d-aware generative model.
\newblock In \emph{Proceedings of the IEEE/CVF Conference on Computer Vision
  and Pattern Recognition}, pages 18440--18449, 2022.

\bibitem[Yang et~al.(2022)Yang, Li, Wu, and Dai]{yang20223dhumangan}
Z.~Yang, S.~Li, W.~Wu, and B.~Dai.
\newblock 3dhumangan: Towards photo-realistic 3d-aware human image generation.
\newblock \emph{arXiv preprint}, arXiv:2212.07378, 2022.

\bibitem[Zhang et~al.(2022{\natexlab{a}})Zhang, Sangineto, Tang, Siarohin,
  Zhong, Sebe, and Wang]{zhang20223d}
J.~Zhang, E.~Sangineto, H.~Tang, A.~Siarohin, Z.~Zhong, N.~Sebe, and W.~Wang.
\newblock 3d-aware semantic-guided generative model for human synthesis.
\newblock In \emph{ECCV}, pages 339--356. Springer, 2022{\natexlab{a}}.

\bibitem[Zhang et~al.(2023{\natexlab{a}})Zhang, Jiang, Yang, Xu, Shi, Song, Xu,
  Wang, and Feng]{Avatargen2023}
J.~Zhang, Z.~Jiang, D.~Yang, H.~Xu, Y.~Shi, G.~Song, Z.~Xu, X.~Wang, and
  J.~Feng.
\newblock Avatargen: A 3d generative model for animatable human avatars.
\newblock \emph{ArXiv}, 2023{\natexlab{a}}.

\bibitem[Zhang et~al.(2023{\natexlab{b}})Zhang, Qiu, Lin, Zhang, Shi, Yang,
  Shi, Yang, Xu, and Yu]{zhang2023dreamface}
L.~Zhang, Q.~Qiu, H.~Lin, Q.~Zhang, C.~Shi, W.~Yang, Y.~Shi, S.~Yang, L.~Xu,
  and J.~Yu.
\newblock Dreamface: Progressive generation of animatable 3d faces under text
  guidance.
\newblock \emph{arXiv preprint arXiv:2304.03117}, 2023{\natexlab{b}}.

\bibitem[Zhang et~al.(2022{\natexlab{b}})Zhang, Zheng, Gao, Zhang, Pan, and
  Yang]{zhang2022multi}
X.~Zhang, Z.~Zheng, D.~Gao, B.~Zhang, P.~Pan, and Y.~Yang.
\newblock Multi-view consistent generative adversarial networks for 3d-aware
  image synthesis.
\newblock In \emph{CVPR}, pages 18450--18459, 2022{\natexlab{b}}.

\bibitem[Zhao et~al.(2022)Zhao, Ma, G{\"u}era, Ren, Schwing, and
  Colburn]{zhao2022generative}
X.~Zhao, F.~Ma, D.~G{\"u}era, Z.~Ren, A.~G. Schwing, and A.~Colburn.
\newblock Generative multiplane images: Making a 2d gan 3d-aware.
\newblock In \emph{ECCV}, pages 18--35. Springer, 2022.

\bibitem[Zhou et~al.(2021)Zhou, Xie, Ni, and Tian]{cips3d}
P.~Zhou, L.~Xie, B.~Ni, and Q.~Tian.
\newblock Cips-3d: A 3d-aware generator of gans based on
  conditionally-independent pixel synthesis.
\newblock \emph{arXiv preprint arXiv:2110.09788}, 2021.

\bibitem[Zhou et~al.(2018)Zhou, Tucker, Flynn, Fyffe, and
  Snavely]{zhou2018stereo}
T.~Zhou, R.~Tucker, J.~Flynn, G.~Fyffe, and N.~Snavely.
\newblock Stereo magnification: Learning view synthesis using multiplane
  images.
\newblock \emph{ACM. Trans. Graph. (SIGGRAPH)}, 2018.

\bibitem[Zhu et~al.(2018)Zhu, Zhang, Zhang, Wu, Torralba, Tenenbaum, and
  Freeman]{VON}
J.-Y. Zhu, Z.~Zhang, C.~Zhang, J.~Wu, A.~Torralba, J.~B. Tenenbaum, and W.~T.
  Freeman.
\newblock Visual object networks: image generation with disentangled 3{D}
  representations.
\newblock In \emph{NeurIPS}, 2018.

\end{thebibliography}
}

\appendix
\newcommand{\AppendixPrefix}{A}
\renewcommand{\thefigure}{\AppendixPrefix\arabic{figure}}
\setcounter{figure}{0}
\renewcommand{\thetable}{\AppendixPrefix\arabic{table}} 
\setcounter{table}{0}
\renewcommand{\theequation}{\AppendixPrefix\arabic{equation}} 
\setcounter{equation}{0}

\section{Single Scene Overfitting}

In addition to the layer ablation experiment described in the main paper for the GAN setting, we also perform a single-scene overfitting experiment. 
We use a textured model downloaded from SketchFab and rendered 400 360-degree views of the model. 
We use 300 views as training data and the remaining images as testing data. 
We appliy the same training settings as described in the main paper and present our method's results with varying numbers of layers.
The last two columns in the presented table show the results of InstantNGP and the ground truth.
To create the mesh layers, we first fitted a SMPL model. However, as the ground-truth mesh had a rather cartoonish body proportion, we manually refined the fitted SMPL model in Blender to approximately match the ground-truth mesh.

\begin{figure}[h]
\centering
\renewcommand{\arraystretch}{0.1}
\setlength{\tabcolsep}{0pt}
\tiny
\begin{tabular}{*{6}{C{0.165\linewidth}}}
1 shell PSNR=22.80 & 4 shells PSNR=24.51 & 12 shells PSNR=27.04 & 24 shells PSNR=27.20 & InstantNGP PSNR=36.86 & GT \\
\includegraphics[width=\linewidth, trim={5cm 0 5cm 7cm}, clip]{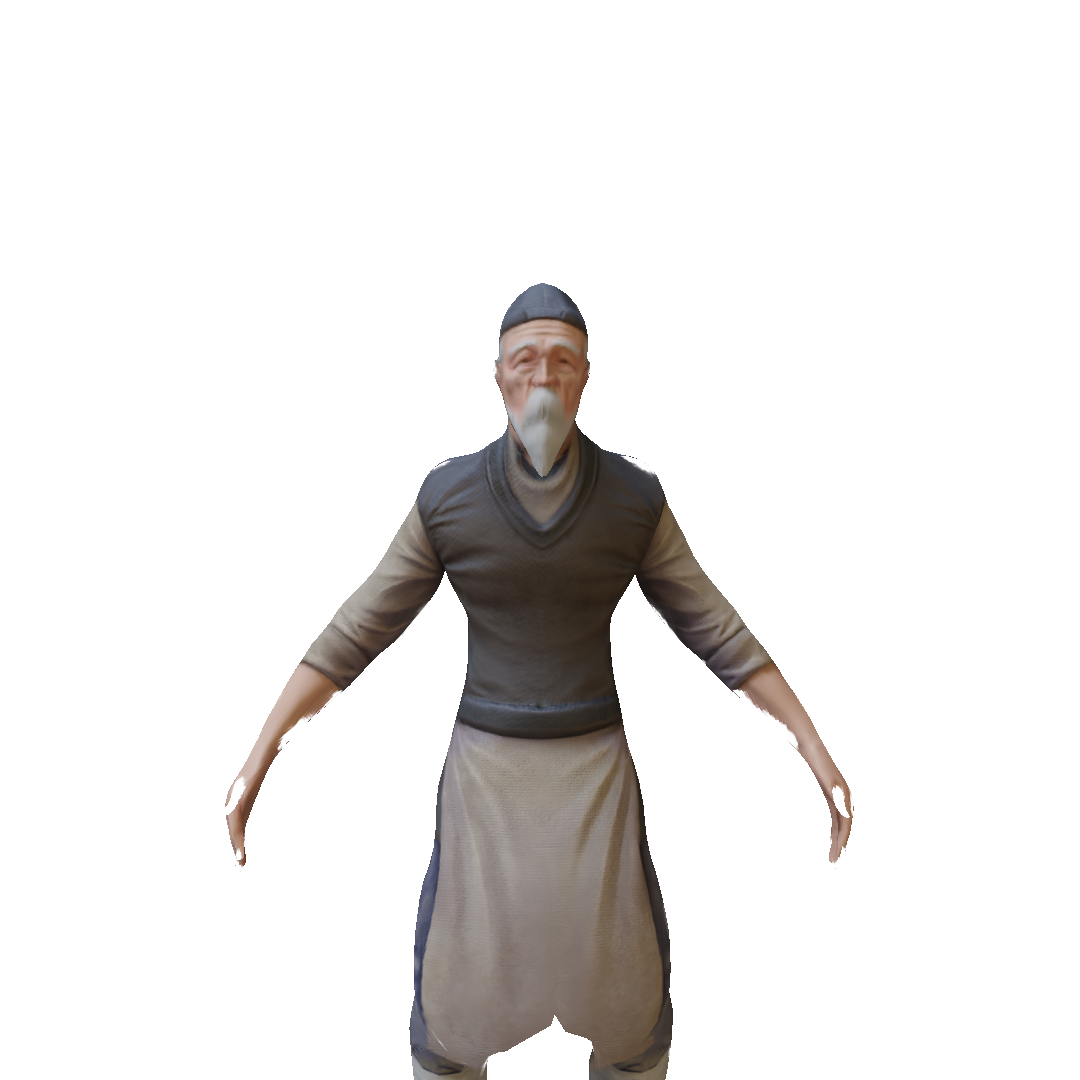} &
\includegraphics[width=\linewidth, trim={5cm 0 5cm 7cm}, clip]{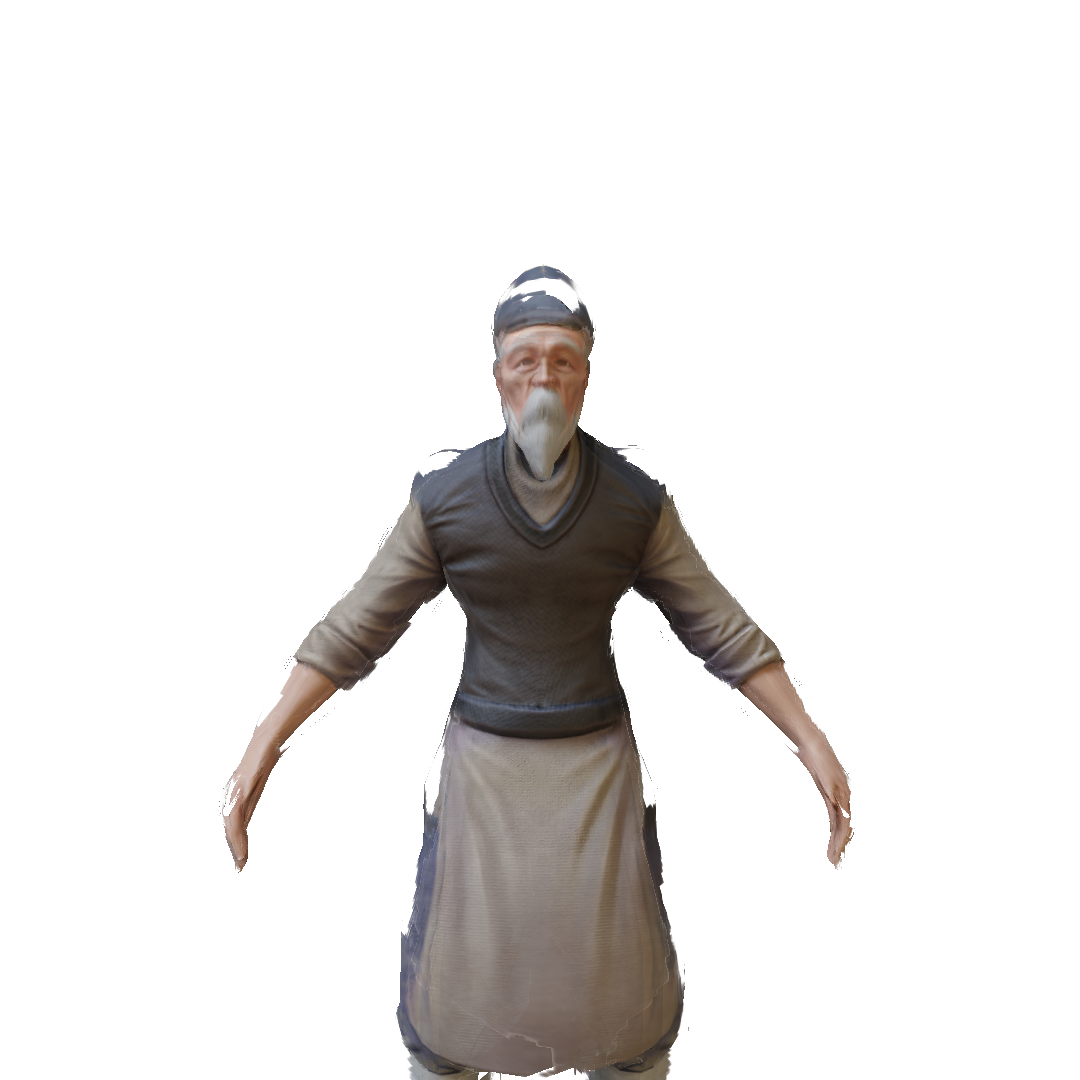} &
\includegraphics[width=\linewidth, trim={5cm 0 5cm 7cm}, clip]{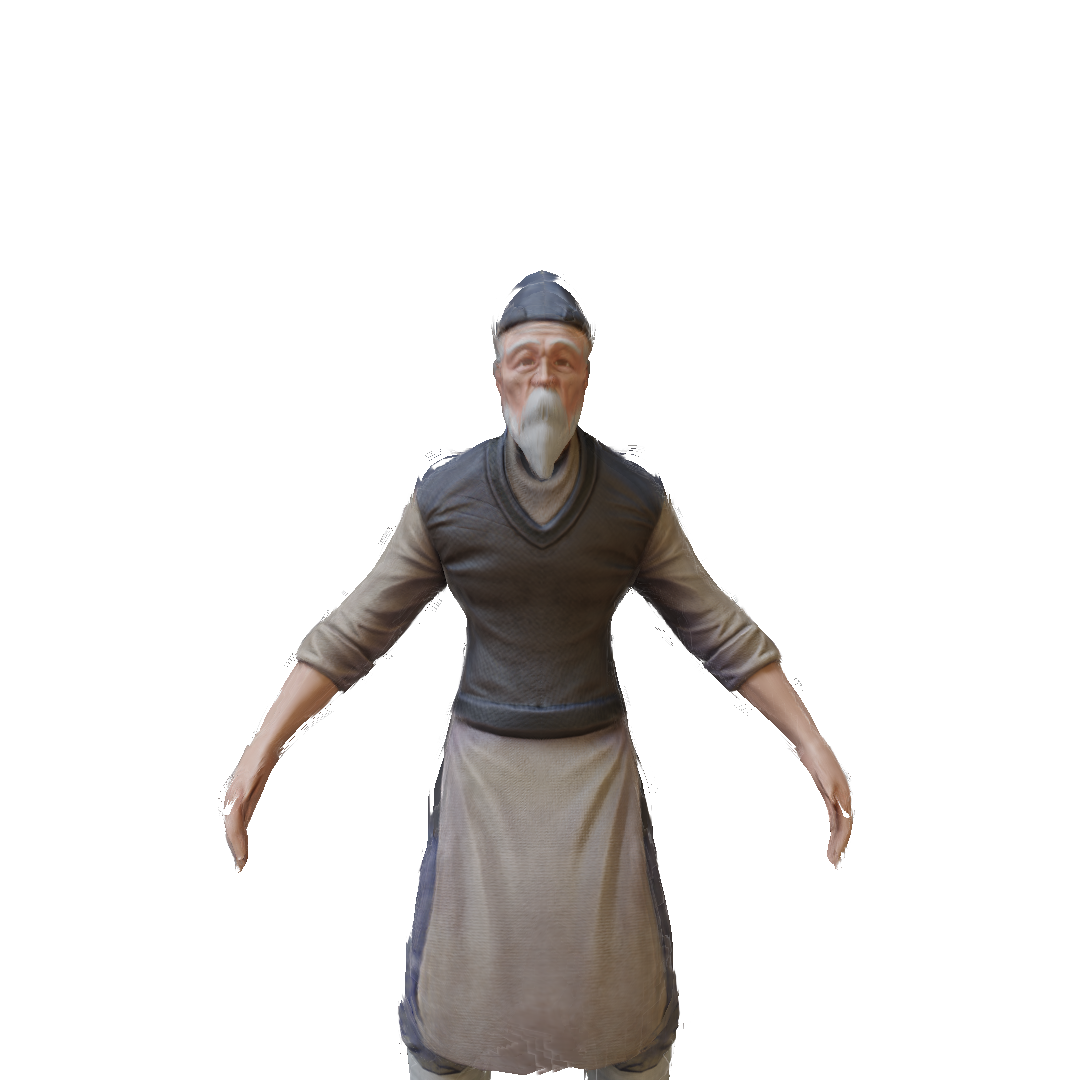} &
\includegraphics[width=\linewidth, trim={5cm 0 5cm 7cm}, clip]{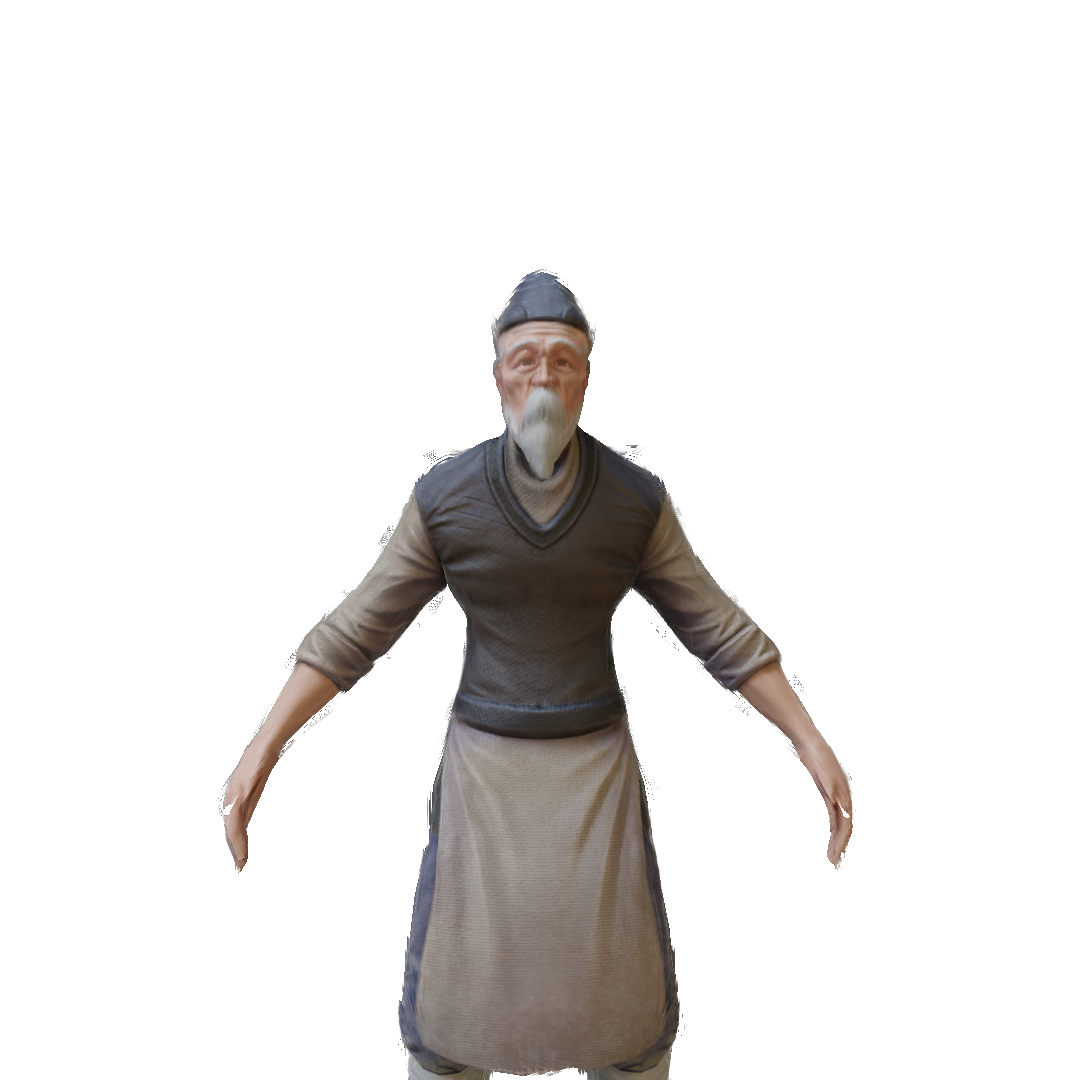} &
\includegraphics[width=\linewidth, trim={5cm 0 5cm 7cm}, clip]{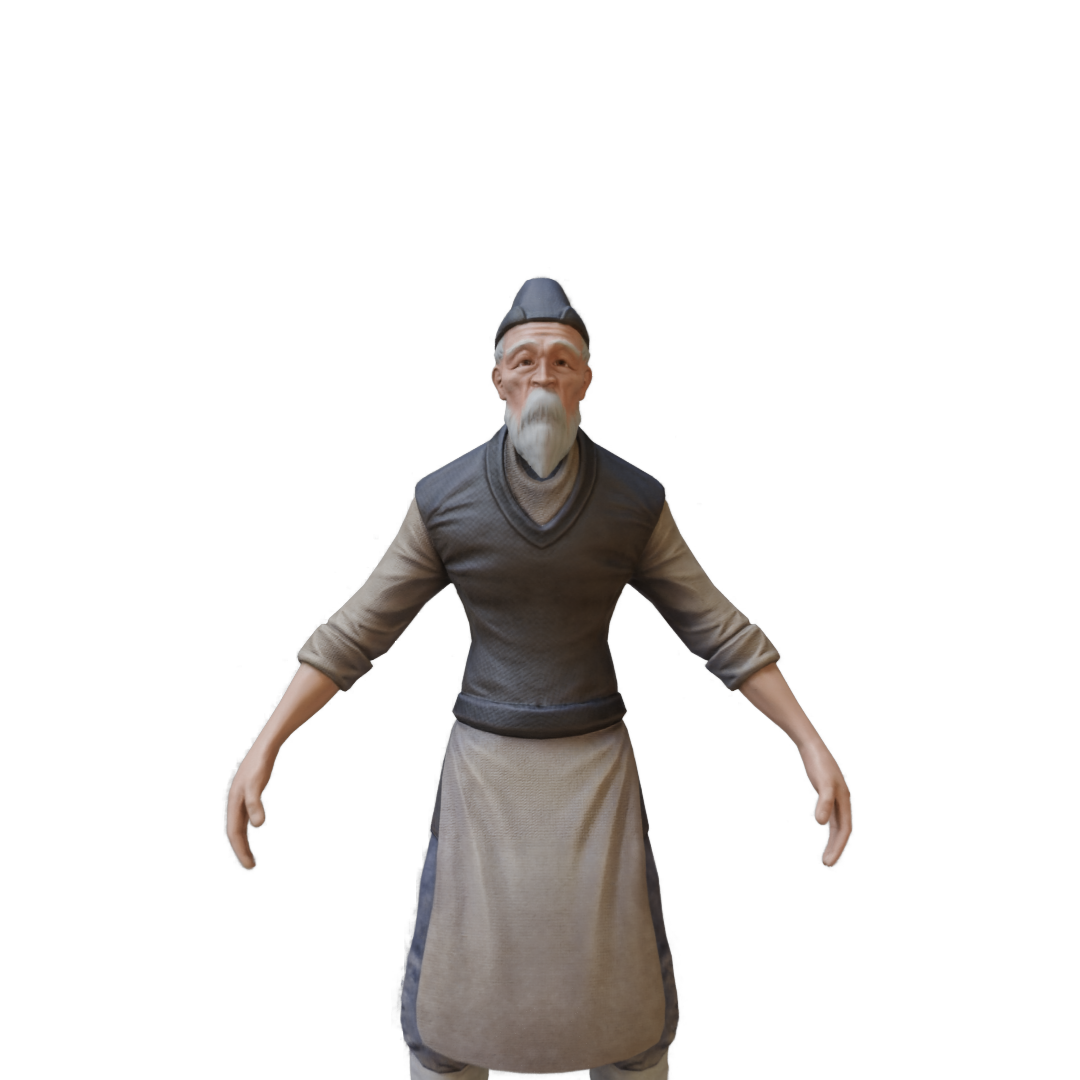} &
\includegraphics[width=\linewidth, trim={5cm 0 5cm 7cm}, clip]{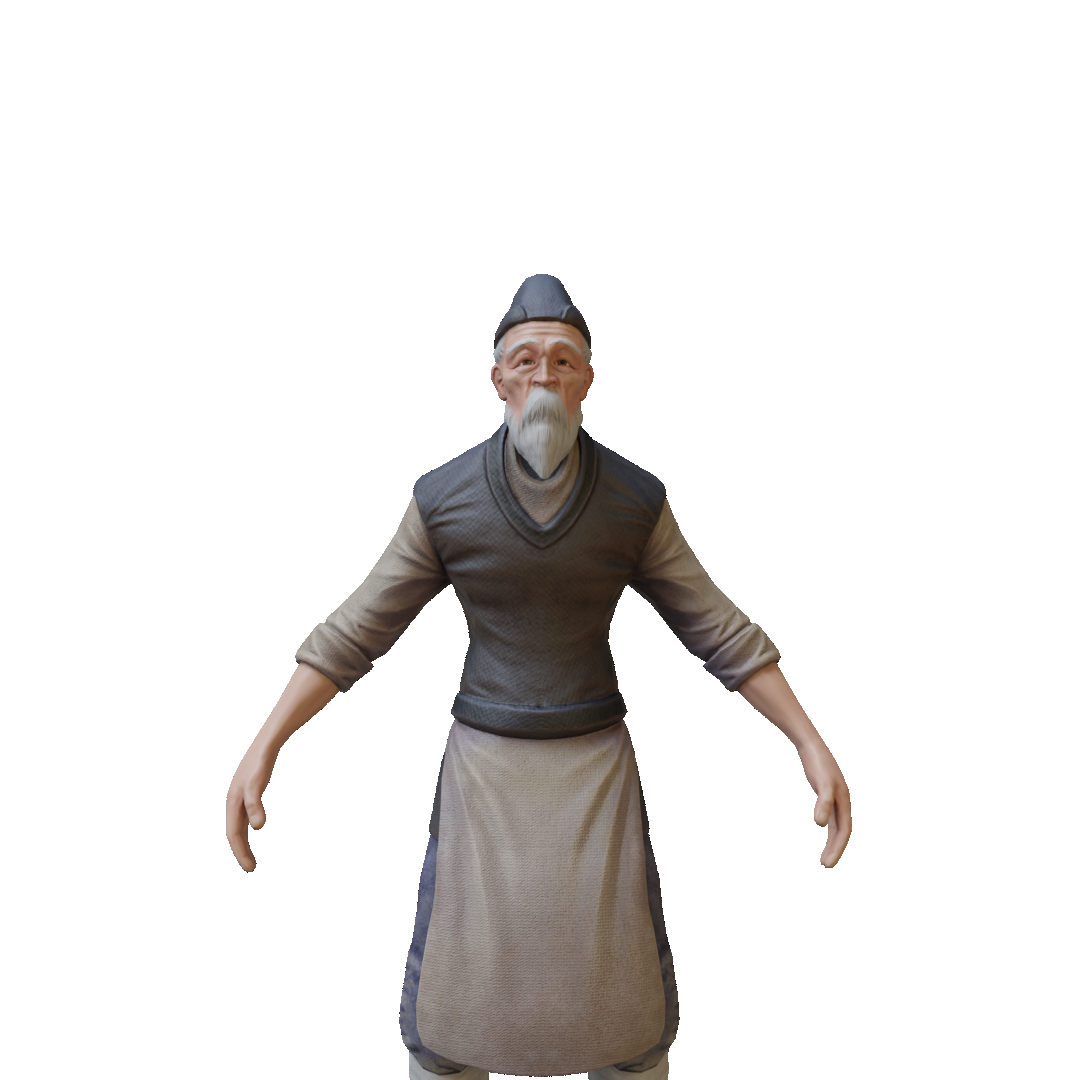} \\
\end{tabular}
\label{fig:sso_result}
\caption{Single-scene overfitting result. We show the results of our method with different number of layers. The last two columns are the results of InstantNGP and ground truth.}
\end{figure}

\section{Implementation Details}

\subsection{Generator}
Our approach employs the generator architecture of StyleGAN2~\cite{stylegan2}, which consists of two components: a mapping network and a convolutional backbone. 
The generator takes a 512-dimensional Gaussian noise input and conditions it using an eight-layer mapping network of 512 hidden units. 
We do not condition the generator on camera pose or body pose. 
The mapping network produces a 512-dimensional latent code, which modulates the layers of the StyleGAN2 convolutional backbone. 
The resulting output is a high-resolution image with 48 channels at $1024\times1024$ resolution. 
To facilitate further processing, we reshape this output into 12 texture planes consisting of RGB and alpha channels, each of shape $1024\times1024\times4$. 
Our architecture is trained from scratch, without using any pretrained networks.

\subsection{Discriminator}

In contrast to EG3D~\cite{eg3d} and GNARF~\cite{bergman2022gnarf}, our framework does not use a dual discriminator because we do not use upsampling to produce the final output images.
Instead, we condition the discriminator on the expected body pose by including the body pose parameters in addition to the camera parameters as input to the mapping network. 
This allows the discriminator to ensure that the applied deformation matches the specified pose. 
To ensure stable training, we add 0.5 standard deviation of Gaussian noise to the body pose parameters before passing them to the discriminator. 
This prevents the discriminator from overfitting to specific poses and cameras in the ground truth data.

\subsection{Training Details}
We defaultly use 12 layers of the surface volume for generation and and the deformation scale is ranges from 0 to 0.05.
We use the Adam optimizer for both the generator and discriminator during optimization, with a learning rate of $2.5 \times 10^{-3}$ for the generator and $2 \times 10^{-3}$ for the discriminator. 

During training, we set the loss weight for R1 regularization to 5 to penalize the gradients of the discriminator and the loss weight of hand regularizer is empirically set to 1.
For the face discriminator, we pad the cropped face into a square shape and resize it to $80\times80$ resolution.

Our models are trained for 4 days on 4 NVIDIA A6000 GPUs, with a batch size of 32. At test time, our model runs at 36 FPS on one NVIDIA A6000 GPU.

\section{Additional Results}

\begin{figure}[t]
\centering
\includegraphics[width=\textwidth]{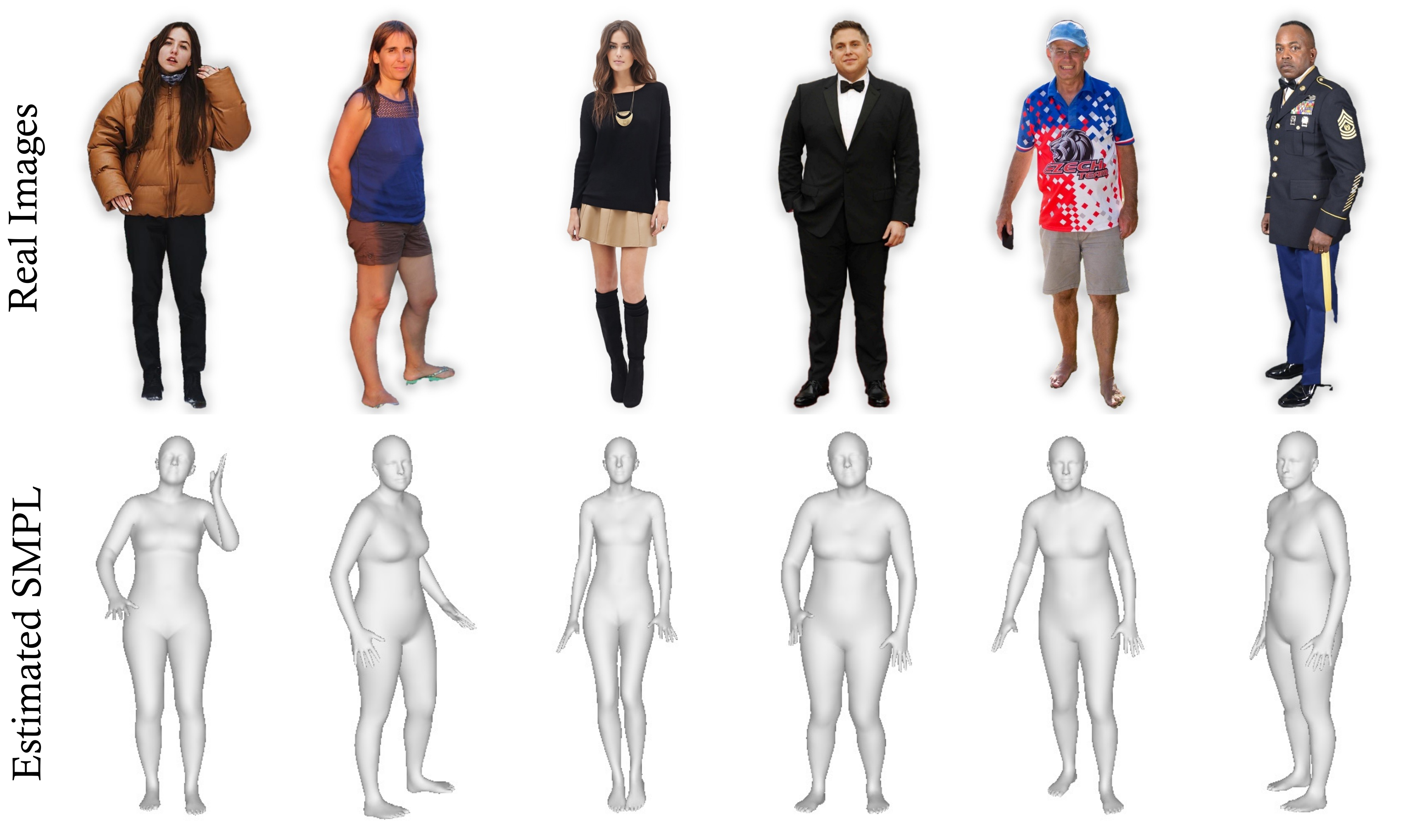}
\caption{Visualization of real images and corresponding estimated SMPL mesh.}
\label{fig:smpl}
\end{figure}

\paragraph{Hand Distribution.}

As shown in \cref{fig:smpl}, we observed that most fingers have a natural curve in dataset, but the SMPL model itself is unable to represent this pose. Most SMPL hand poses are naturally open, and as a result, even with deformation, the hand mesh cannot accurately fit the hand pose distribution of dataset. This can lead to some fingers appearing translucent. To address this issue, we reduce the original deformation scale and aim to learn an opaque texture to ensure the photorealistic appearance of the hand. In future work, we plan to use a more precise SMPL-H model to further improve the representation of the hand pose.


\begin{figure}[t]
\centering
\includegraphics[width=\textwidth]{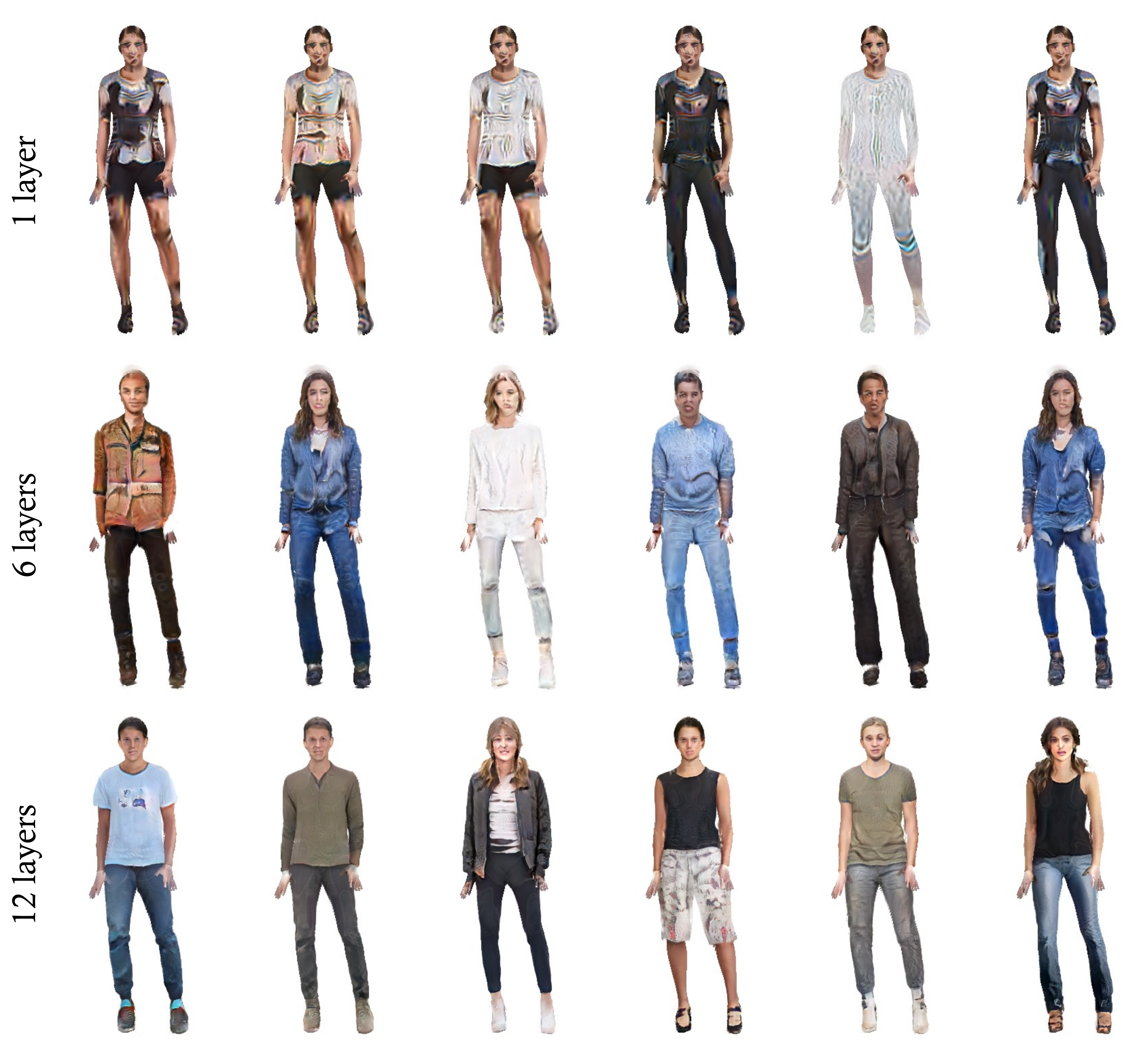}
\caption{Qualitative comparison for ablations on number of layers.}
\label{fig:layers}
\end{figure}

\paragraph{Number of Layers.}
In \cref{fig:layers}, we present visualizations of the generated humans at different numbers of layers. As the number of layers increases, we observe significant improvements in the quality and diversity of both the images and faces generated by our model.
In the case with only one layer, we rasterize the texture onto the original SMPL surface, resulting in very thin people and limited ability to synthesize realistic human images. This approach struggles to handle complex hair and clothing structures.
However, as the number of layers increases, the layered surface volumes can capture more details and volumetric structures, such as hair and clothing, resulting in more realistic and diverse human images.

\end{document}